\documentclass[review]{elsarticle} 
\usepackage{graphicx} 
\usepackage[left]{lineno} 
\usepackage[a4paper, margin=1in]{geometry} 
\usepackage{url}
\usepackage{appendix}
\usepackage{breqn}
\usepackage{breqn}
\usepackage{xcolor}
\usepackage{amssymb}     
\usepackage{amsfonts}    
\usepackage{mathtools}   
\usepackage{bm}          
\usepackage{algorithm}        
\usepackage{algorithmic}      
\usepackage{listings}         
\usepackage{booktabs}    
\usepackage{array}       
\usepackage{multirow}    
\usepackage{tabularx}    
\usepackage{textcomp}

\usepackage{graphicx} 
\usepackage[left]{lineno} 
\usepackage[a4paper, margin=1in]{geometry} 
\usepackage{amsmath}     
\usepackage{amssymb}     
\usepackage{amsfonts}    
\usepackage{mathtools}   
\usepackage{bm}          
\usepackage{algorithm}        
\usepackage{algorithmic}      
\usepackage{listings}         
\usepackage{booktabs}    
\usepackage{array}       
\usepackage{multirow}    
\usepackage{tabularx}    
\usepackage{setspace}    

\usepackage{natbib}  

\date{} 

\begin{document}

\begin{flushleft}
    \Large \textbf{Multi-class Seismic Building Damage Assessment from InSAR Imagery using Quadratic Variational Causal Bayesian Inference} \\
    \vspace{0.5cm}
    \normalsize
    Xuechun Li\textsuperscript{1}, Susu Xu\textsuperscript{1,2,*}\\
    \vspace{0.3cm}
    \textsuperscript{1}\textit{Center for Systems Science and Engineering, Department of Civil and Systems Engineering, Johns Hopkins University}\\
    \textsuperscript{2}\textit{Data Science and AI Institute, Johns Hopkins University}\\
    \textsuperscript{*}\textit{Corresponding author: Susu Xu, email: \url{susuxu@jhu.edu}}\footnote{201P Latrobe Hall, 3400 North Charles Street, Baltimore, MD 21218}
\end{flushleft}

\noindent 
\section*{Abstract}
Interferometric Synthetic Aperture Radar (InSAR) technology leverages satellite radar signals to detect surface deformation patterns, allowing large-scale monitoring of earthquake impacts on the built environment. These measurements can reveal structural responses of buildings to seismic forces, making InSAR valuable for rapid damage assessment. Although detailed multi-class building damage classifications (e.g., no damage, moderate, severe, collapse) are essential for emergency response planning during the critical first 72 hours, extracting this information from InSAR data presents significant challenges: the overlapping of building damage signatures with secondary hazards and environmental noise, computational intractability due to the exponential growth in model parameters in multi-class scenarios, and the need for rapid processing at regional scales. We address these challenges through a novel multi-class variational causal Bayesian inference framework with quadratic variational bounds - an approach that provides mathematically rigorous approximations of complex damage patterns while ensuring computational efficiency. By integrating InSAR observations with USGS ground failure models and building fragility functions, our framework effectively separates building damage signals while maintaining computational efficiency through strategic pruning. Evaluation across five major earthquakes (Haiti 2021, Puerto Rico 2020, Zagreb 2020, Italy 2016, Ridgecrest 2019) demonstrates substantial improvements in damage classification accuracy (AUC: 0.94-0.96), achieving up to 35.7\% improvement over existing approaches. Our method shows particular robustness in areas with multiple overlapping hazard signatures, maintaining high accuracy (AUC $\geq$ 0.93) across all damage categories while reducing computational overhead by over 40\% through strategic pruning, all without requiring extensive ground truth data.

\noindent \textit{\textbf{Keywords:} InSAR, earthquake damage assessment, quadratic variational bounds, causal Bayesian inference, multi-hazard analysis, building damage classification, remote sensing, disaster response }

\section{Introduction}
Natural disasters, particularly earthquakes, result in significant human and economic losses, largely due to building damage. The first 72 hours after an earthquake—known as the "Golden Hours"—are critical for saving lives \cite{li2023disasternet}. During this period, emergency responders must rapidly rescue survivors, assess facility safety, and relocate evacuees. These operations depend on detailed understanding of building damage patterns. Multi-class building damage classification is essential for optimizing response strategies—from prioritizing rescue operations in severely damaged areas to efficiently allocating repair resources in moderately affected zones. Studies show that reducing assessment time by 6-12 hours can significantly improve rescue success rates, emphasizing the need for rapid and reliable damage classification methods \cite{erdik2011rapid}.

However, rapid damage assessment faces significant technical challenges due to the complex nature of earthquake events. Earthquakes trigger multiple concurrent hazards beyond direct structural damage, including landslides and soil liquefaction. These secondary hazards often co-occur and interact with building damage, creating complex patterns that can mask or amplify each other. This interplay of multiple hazards complicates the accurate assessment of building damage levels.

Post-disaster damage assessment methods have evolved substantially over recent decades. Traditional approaches relied heavily on manual inspection - a process that proved labor-intensive, costly, and logistically challenging in large-scale disasters ~\cite{gueguen2015large}. The 2021 Haiti earthquake illustrates these limitations: when over 52,000 buildings were damaged or destroyed, engineering teams from multiple humanitarian organizations required several weeks to months to complete their ground assessments \cite{undp2021haiti}. Various approaches have been developed to address these limitations, but each faces significant challenges. Statistical methods attempted to combine historical inventory data with geospatial proxies (e.g., slope, lithology) ~\cite{marc2016seismologically, nowicki2018global, zhu2017updated}, but these models often struggle with complex overlapping disaster events and insufficient building typology data. The limited availability of geospatial proxy layers and the uncertainties of single-hazard modeling often constrain their resolution and accuracy. Moreover, since hazards and impact patterns are sensitive to subtle environmental and geological factors that vary from region to region, adapting and generalizing statistical models trained on past events to new events remains challenging.

Recent advances in remote sensing technologies, particularly Interferometric Synthetic Aperture Radar (InSAR), have transformed post-disaster damage assessment capabilities. InSAR technology uses satellite radar signals to measure land-surface deformation with high precision, enabling rapid detection of structural damage through surface deformation patterns \cite{yun2015rapid}. The advancement of such sensing technologies has made large-scale observation data widely available for estimating disaster-induced multi-hazards and impacts. The integration of multiple remote sensing technologies has further enhanced our ability to assess multi-hazard scenarios. For instance, combined analysis of InSAR and optical satellite imagery has improved detection of both structural damage and secondary hazards \cite{adriano2019multi, mondini2021landslide}. Beyond earthquakes \cite{yun2015damage}, satellite images can now provide information on disaster-induced ground surface changes \cite{barras2007satellite} within hours or days after disasters. Additional data sources, such as social media, can provide near-immediate information about societal impacts \cite{yates2011emergency, wang2024near}.

However, these modern sensing approaches face their own set of challenges. Although some studies have attempted to establish quantitative relationships between InSAR-derived surface deformation and building damage levels, they face significant challenges in distinguishing building damage from secondary hazards such as landslides and liquefaction that often co-occur at the same locations~\cite{li2021exploring} or in performing multi-class damage assessment~\cite{xu2022deep, xu2022seismic, wang2023causality, li2023disasternet, li2024rapid, yu2024intelligent, li2024spatial, wang2024scalable}. It is particularly challenging to directly categorize different changes from these sensing observations when ground failures, building damage, and noise from vegetation growth and anthropogenic modifications are co-located \cite{yun2015rapid}. While some existing approaches use linear combinations or black-box supervised classifiers to incorporate geospatial features and sensing observations for single-type hazard and impact estimation, these models have a fundamental limitation. They fail to capture the chain of events in earthquake scenarios - how ground shaking triggers both direct building damage and secondary hazards like landslides, which can then cause additional damage. This inability to model such complex relationships limits their applicability to common multi-hazard scenarios. Previous approaches have achieved AUC scores ranging from 0.82 to 0.89 for binary damage classification \cite{rao2023earthquake, xu2022seismic}, but multi-class assessment has remained elusive.

Current methods face a fundamental trade-off between computational efficiency and classification accuracy. Linear models process data quickly but oversimplify damage patterns, while complex neural networks achieve higher accuracy but are too slow for emergency response timeframes \cite{xu2022seismic}. A mathematical framework that balances these competing demands is urgently needed. In addressing post-disaster building damage assessment using InSAR technology, our research tackles three critical challenges. First, the interpretation of InSAR measurements is complicated by multiple overlapping signals: building damage patterns are often masked by both secondary hazards (such as landslides and liquefaction) and environmental noise (including atmospheric disturbances, topographical variations, and urban clutter), making signal separation and damage assessment particularly challenging. Second, implementing multi-class damage assessment at scale introduces significant computational complexity due to parameter explosion in the model space, especially when considering multiple damage states and their interactions with various hazards. Third, processing large-scale post-disaster data requires efficient computational strategies, particularly within the critical response window.

\begin{figure}
\centering
\includegraphics[width=0.8\textwidth]{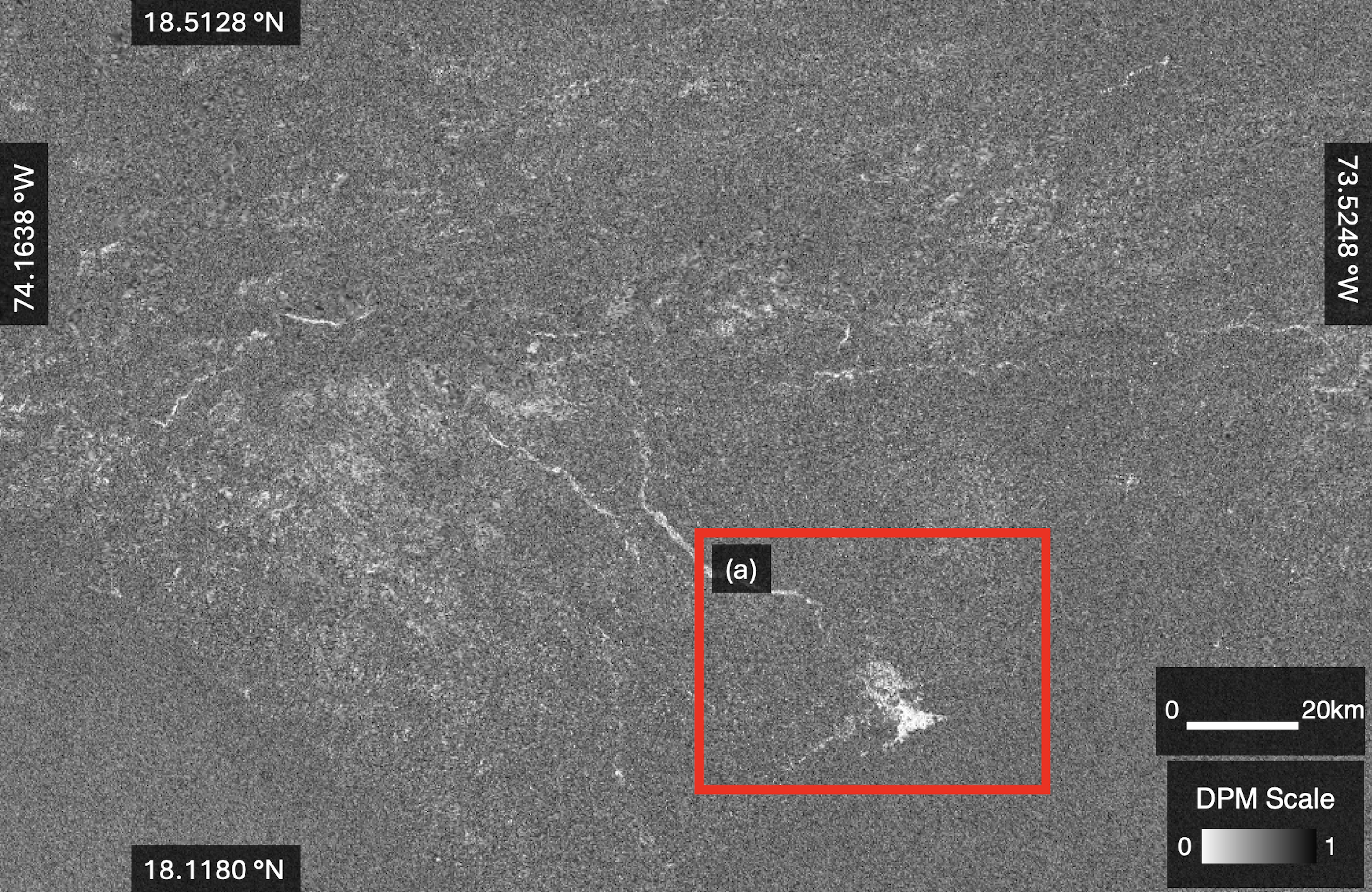}
\caption{\textbf{Damage proxy map (DPM) derived from Interferometric Synthetic Aperture Radar (InSAR) data by NASA's Advanced Rapid Imaging and Analysis (ARIA) team \cite{ARIA} after the 2021 Haiti earthquake.} The map reveals surface deformation patterns by comparing pre- and post-earthquake satellite radar measurements. The normalized color scale (0-1) quantifies surface changes, where whiter areas (values closer to 1) indicate significant surface alterations typically associated with severe structural damage, landslides, or major ground deformation, while darker areas (values closer to 0) suggest minimal change to the ground surface. This map demonstrates the complex spatial distribution of earthquake impacts across both densely built urban areas and steep mountainous terrain, illustrating the challenge of distinguishing between different types of surface changes. The red box highlights an area of particular interest where multiple types of surface changes overlap.}
\label{DPM_example}
\end{figure}

To address these challenges, we introduce a novel quadratic variational causal Bayesian (QVCBI) framework that enhances InSAR technology for earthquake-induced building damage assessment. Our approach provides two key advantages over existing methods: (1) quadratic variational bounds that capture non-linear relationships between damage patterns with provably tighter approximations than linear or first-order methods \cite{hoffman2013stochastic}, and (2) computational tractability through exploitation of sparse causal dependencies \cite{spiegelhalter1998bayesian}, enabling processing of regional-scale disasters within 4 hours. By integrating InSAR observations with U.S. Geological Survey (USGS) ground failure models and fragility functions, our framework effectively separates building damage signals from overlapping hazard and environmental noise. To enable rapid processing of large-scale disasters, we implement a local pruning strategy that effectively reduces complexity in causal graph processing, making our model particularly suitable for time-critical disaster response across extensive geographical areas.

QVCBI makes three primary contributions to the field of InSAR-based damage assessment. First, we introduce an expressive causal Bayesian network that advances beyond traditional binary classification to handle multi-class damage assessment. This network enables joint modeling and inference of multiple damage classes and seismic-induced hazards while capturing their complex causal dependencies, achieving high-resolution estimation of cascading hazards in a physically interpretable manner. Second, we develop a unified approach to handling both overlapping hazard signatures and environmental noise by formulating them as interrelated random variables within our Bayesian framework, enabling reliable discrimination between actual damage patterns and various sources of interference through systematic uncertainty quantification. Third, we derive and implement quadratic variational bounds that effectively capture non-linear relationships while controlling computational complexity, enabling accurate approximation of complex, multi-modal posterior distributions needed for multi-class damage classification while ensuring computational efficiency through tight bounds and strategic pruning. These innovations make QVCBI practical for time-critical disaster response applications where both accuracy and rapid processing are essential.

\begin{figure}[t]
    \centering
\includegraphics[width=0.5\textwidth]{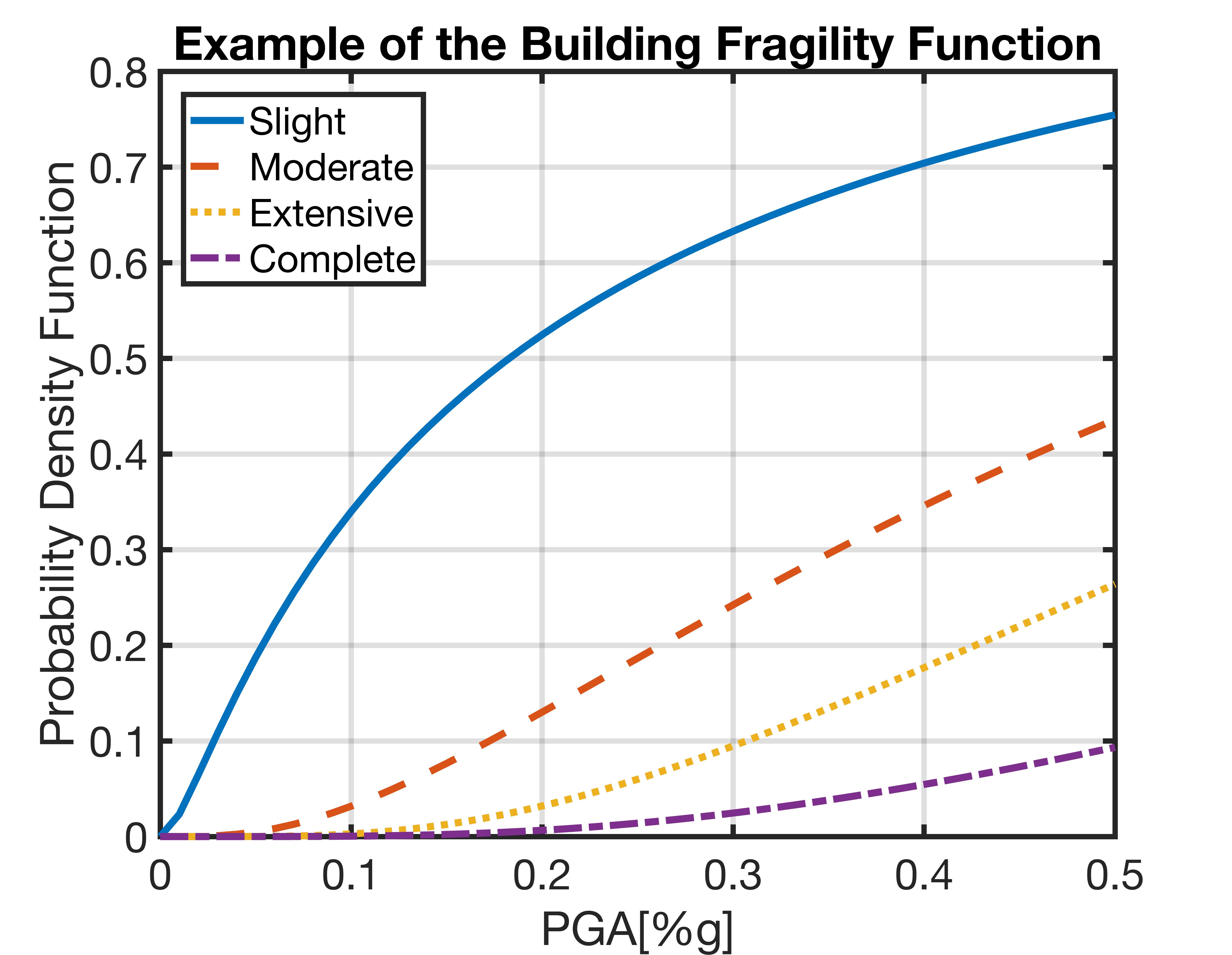}
    \caption{HAZUS Building fragility curves for multi-categorical building damage.}
    \label{Fragility}
\end{figure}

\section{Study cases and datasets}

We evaluate the performance of our model using five earthquakes: the 2021 Haiti earthquake, the 2020 Puerto Rico earthquake, the 2020 Zagreb earthquake, the 2016 Italy earthquake, and the 2019 Ridgecrest earthquake. We utilized Damage Proxy Maps (DPMs), which are derived from comparing pre- and post-earthquake InSAR measurements, to analyze these earthquakes. All the experiments utilize the peak ground acceleration (PGA) data from the ShakeMap \cite{worden2020shakemap}, published by the USGS, the ground failure models from the USGS ground failure model \cite{zhu2017updated,nowicki2018global}, and the building footprint (BF) maps obtained from the OpenStreetMap (OSM) project \cite{OpenStreetMap}. In our experiments, we utilize different building fragility functions \cite{fema2003hazus, wald2010pager} to evaluate the performance of our posteriors for building damage. A brief overview of the datasets we used in our experiments is shown in the following sections.

\subsection{InSAR based building damage estimation}

\begin{figure}[t]
    \centering
\includegraphics[width=0.8\textwidth]{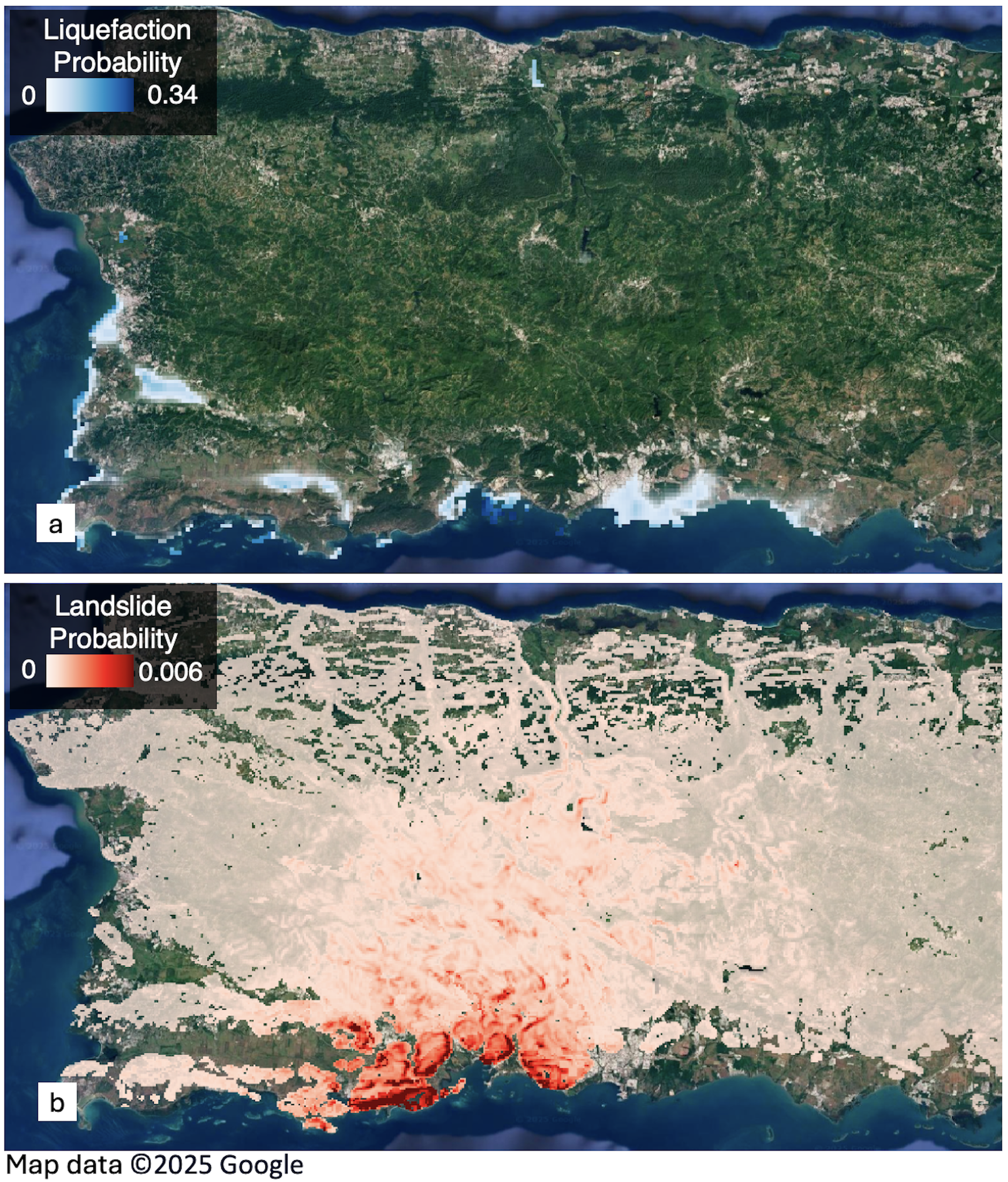}
    \caption{Example of ground failure models for landslide and liquefaction produced by the USGS after the 2020 Puerto Rico earthquake sequence. The legend colors represent the probability of ground failure models.}
    \label{USGS_GF}
\end{figure}

Over recent years, numerous models have been developed to estimate multiple hazards and impacts for post-event situational awareness. Among various remote sensing technologies, InSAR has emerged as a particularly promising method for estimating building damage, showing the potential to provide accurate and timely information after natural disasters. InSAR technology employs two SAR images of the same area captured at different times to create an interferogram, revealing changes in the surface of the earth. A significant advancement in InSAR-based damage assessment came from NASA's Advanced Rapid Imaging and Analysis (ARIA) team through their development of Damage Proxy Maps (DPMs). These maps are generated using Synthetic Aperture Radar (SAR) data from the Copernicus Sentinel-1A and Sentinel-1B satellites \cite{dpm1}, analyzing coherence changes between pre- and post-event images to identify potential structural damage. Figure \ref{DPM_example} shows an example DPM where the brightness scale (0-1) indicates the degree of surface change, with brighter areas suggesting more significant modifications to the surface structure, potentially indicating severe damage or ground failures.

When an earthquake occurs, it causes ground shaking and geospatial feature changes, resulting in building damage, landslides, and liquefaction that induce surface ground change signals detectable in InSAR measurements and DPMs, as illustrated by the bright patterns in Figure \ref{DPM_example}(a). However, directly categorizing different types of changes using these data can be challenging due to the common overlap and co-location of building damage, ground failures, noise from vegetation growth, and anthropogenic modifications. While InSAR-based methods show promise, they face limitations such as atmospheric disturbances, topographical effects, and reduced accuracy in areas with dense vegetation \cite{havivi2018combining}. Consequently, these methods need to be enhanced with additional approaches to ensure precise and comprehensive damage assessment.

\subsection{Building Fragility Curves and PAGER System}

Building damage estimation methods commonly rely on fragility curves - statistical functions that estimate the probability of varying degrees of damage based on ground motion intensity measures, such as PGA or spectral acceleration. These curves help represent earthquake vulnerability across different damage levels, ranging from no damage to complete destruction \cite{bozorgnia2004earthquake, biglari2020damage}. The curves are typically expressed as log-normal functions characterized by two parameters (log-median and log-standard deviation) and are developed as a function of ground motion intensity.

In QVCBI, we specifically utilize HAZUS fragility curves \cite{fema2003hazus}, which are widely adopted in the United States and many other regions. HAZUS provides standardized fragility functions that describe the probability of reaching or exceeding structural and nonstructural damage states, given median estimates of spectral response \cite{kircher1997development}. These curves distribute damage among four states: slight, moderate, extensive, and complete, with discrete damage-state probabilities calculated as the difference between cumulative probabilities of reaching successive damage states. At any given response level, the probabilities of a building reaching various damage levels sum to 100\% \cite{fragility}. Figure \ref{Fragility} shows an example of the HAZUS fragility function for multi-categorical building damage. However, HAZUS fragility curves face several limitations. They are developed based on empirical data from specific regions and may not fully capture all sources of uncertainty and variability, including site-specific conditions, soil-structure interaction, and various earthquake types \cite{baker2011limitations}. Additionally, these curves might not be applicable to structures significantly different from those used in their development, particularly in regions with unique building types and construction practices \cite{bommer2009fragility}. 

To address these limitations, we also incorporate the Prompt Assessment of Global Earthquakes for Response (PAGER) system, an automated system developed by the U.S. Geological Survey that assesses earthquake impacts worldwide \cite{pager1}. PAGER provides rapid estimates of fatality and economic losses, informing emergency responders, government agencies, and media about potential disaster scope \cite{pager2}. The system calculates ground shaking estimates using ShakeMap methodology \cite{pager3} and incorporates peak ground velocity (PGV) data to generate refined fragility models that are more adaptable to different global regions \cite{pager4}.

By utilizing both HAZUS and PAGER in QVCBI, we leverage their complementary strengths: HAZUS provides detailed multi-class damage state definitions and well-calibrated fragility curves for standard building types, while PAGER offers broader global applicability and rapid assessment capabilities. This combination helps QVCBI maintain reliability across diverse geographical regions and building types.

\subsection{Ground failure models}

The USGS ground failure models provide rapid estimates of earthquake-induced landslide \cite{nowicki2018global} and liquefaction \cite{zhu2017updated} probabilities. These models are critical components of earthquake hazard assessment, as they estimate areas susceptible to secondary ground failures that can significantly impact building damage patterns. The models generate probability maps for both landslide and liquefaction hazards shortly after significant earthquakes, as illustrated in Figure \ref{USGS_GF}, which shows the distinct spatial patterns of estimated hazards.
Figure \ref{USGS_GF} demonstrates the different characteristics of these hazards in a coastal region: (a) liquefaction probability map showing higher susceptibility in coastal and low-lying areas with values ranging from 0 to 0.34, and (b) landslide probability map indicating increased risk in areas of high relief and steep slopes, with probabilities ranging from 0 to 0.006. The contrasting spatial patterns reflect the different physical processes and geological conditions that control these hazards.

The USGS liquefaction model considers factors such as soil properties, groundwater conditions, and ground motion intensity to estimate the probability of soil losing its strength during earthquake shaking. The model is particularly sensitive to the presence of saturated, unconsolidated sediments, which explains the higher probabilities often observed in coastal and riverine areas \cite{zhu2017updated}. The landslide probability model, on the other hand, incorporates topographic data, geological conditions, and ground motion parameters to identify areas susceptible to earthquake-induced slope failures. The model accounts for factors such as slope angle, rock strength, and ground acceleration \cite{nowicki2018global}. 

These ground failure models serve as crucial prior information in QVCBI, helping to disambiguate between different sources of surface deformation observed in InSAR measurements. By incorporating these probability maps into our causal Bayesian network, we can better distinguish between building damage and secondary ground failures, ultimately improving the accuracy of our multi-class building damage assessment.

\subsection{Study cases}

\subsubsection{The 2021 Haiti earthquake} On 14 August 2021, an Mw 7.2 earthquake occurred in the southern peninsula of Haiti. At least 2,248 people were killed, 53,815 homes were destroyed, and 83,770 were damaged throughout Grand'Anse, according to the post-disaster reports\cite{Haiti_report}. The prior landslide and liquefaction are provided by USGS \cite{Haiti_LSLF}. The ARIA team generated DPMs using Sentinel-1 SAR images\cite{Haiti_ARIA}. Ground truth inventories for building damage were later collected by GEER team~\cite{HaitiBDGT,HaitiGEER}.
    
\subsubsection{The 2020 Puerto Rico earthquake} The magnitude 6.4 earthquake struck the southwest area of Puerto Rico on Jan. 7, 2020. Post-earthquake reports show that at least 4,893 landslides were triggered across the Tiburon Peninsula by the earthquake and subsequent rainfall from Tropical Cyclone Grace \cite{28}. More than 775 buildings were affected. The ARIA team generated DPMs for this earthquake using the SAR images from the Copernicus Sentinel-1 satellites of the European Space Agency.

\subsubsection{The 2020 Zagreb, Croatia} 
A magnitude 5.3 earthquake struck Zagreb, Croatia, on March 22, 2020\cite{zagreb_info}. Damage reports show that 26,197 buildings were damaged or destroyed, including the historic Zagreb Cathedral.\cite{zagreb_impact}. After the earthquake, the ARIA team provided DPMs for the earthquake using the Sentinel-1 SAR images\cite{zagreb_dpm}. USGS provided the prior landslide and liquefaction estimates generated using the ShakeMap\cite{zagreb_sm}. Ground truth data was collected by the GEER team after the earthquake \cite{GEER2020}.
   
\subsubsection{The 2016 Italy earthquake} The 2016 Mw 6.1 Italy earthquake occurred on October 26, 2016, resulting in several dozen people injured and numerous buildings damaged \cite{Italy_impact}. The ARIA team generated DPMs using the Italian Space Agency’s COSMO-SkyMed satellites and Japan Aerospace Exploration Agency’s ALOS-2 satellites SAR images \cite{Italy_ARIA}. In this experiment, we utilize the HAZUS fragility curve and PAGER to test the model performance under different prior models and choose the one with the best performance. Ground truth inventories for building damage were later collected by GEER team~\cite{HaitiBDGT,HaitiGEER}.

\begin{figure}[h]
\centering
\includegraphics[width=\textwidth]{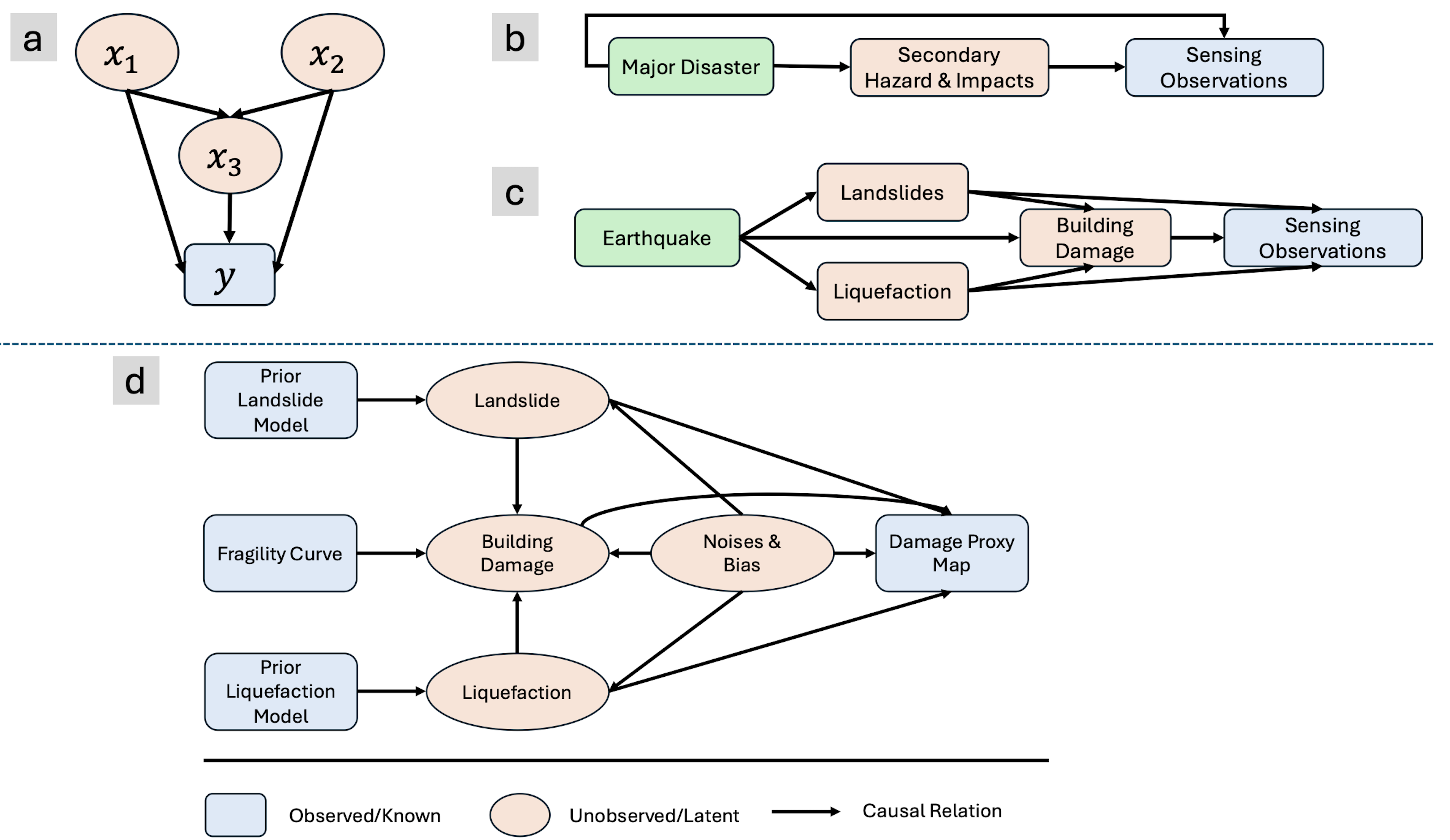}
\caption{\textbf{Multi-level representation of our causal modeling approach.} (a) A simple example of a Bayesian network structure showing variables and their conditional dependencies. (b) High-level conceptual framework illustrating how major disasters trigger secondary hazards, which in turn produce observable impacts through remote sensing. (c) Earthquake-specific instantiation of the disaster chain, demonstrating how seismic events lead to multiple concurrent hazards (landslides, liquefaction, building damage) that generate detectable sensing observations. (d) Our complete QVCBI framework implementation, showing the integration of prior models (blue boxes), unobserved hazard states, and observable measurements, with arrows indicating causal relationships and information flow. The structure explicitly models the interactions between various hazards and their combined effects on damage proxy measurements.}
\label{BN_all}
\end{figure}

\subsubsection{The 2019 Ridgecrest, California earthquake} The Mw 7.1 earthquake in eastern California, southwest of Searles Valley, occurred on July 6th, 2019, at 03:19 UTC. Damage estimated more than \$100 million US dollars\cite{RC_summary}. After the earthquake, the ARIA team provided DPMs using the Sentinel-1 SAR images. Researchers from the USGS, the University of Puerto Rico Mayagüez, the GEER team, and the Structural Extreme Events Reconnaissance team later conducted field reconnaissance to collect ground truth observations \cite{allstadt2022ground,miranda2020,miranda2020b}. To test our model robustness to different prior building fragility functions, we consider three different prior models: (1) without using any prior information for building damage; (2) apply the randomly generated prior information for building damage; (3) utilize the HAZUS building fragility curve.

\section{Background and methodology}

\subsection{Background}
\subsubsection{Bayesian networks}
A Bayesian network (BN) is a directed acyclic graph (DAG) that represents probabilistic relationships among a set of variables. These networks are particularly powerful for modeling complex cause-and-effect relationships, making them ideal for disaster impact assessment where multiple factors interact. As shown in Figure \ref{BN_all}(a), nodes in the network represent variables (such as hazards or observations), while edges represent conditional dependencies between these variables.

The fundamental goal of a Bayesian network is to calculate the conditional posterior probability distribution, $p(\textbf{x}|\textbf{y})$, for unobserved variables $\textbf{x}$ given observed variables $\textbf{y}$. A key property of Bayesian networks is that given its parent nodes, any node is conditionally independent of all non-descendants. This allows the joint probability distribution to factorize into a series of simpler conditional probability distributions, making complex problems computationally tractable \cite{spiegelhalter1998bayesian}.
In our application, we leverage this framework to model the intricate relationships between earthquakes, their induced hazards, and observed damage patterns. As illustrated in Figure \ref{BN_all}(c), we construct a causal network that connects initial earthquakes to secondary hazards (including building damage, landslides, and liquefaction) and ultimately to the observed changes in DPM measurements. This structure allows us to systematically account for how different hazards contribute to the observed surface changes.

\subsubsection{Variational inference} While Bayesian networks provide a powerful framework for modeling relationships between variables, computing exact posterior distributions in complex networks can be computationally intractable. Variational inference offers a practical solution by approximating these complex posterior distributions through optimization, making it particularly suitable for our large-scale multi-hazard assessment problem.

In variational inference, we approximate the true posterior distribution $p(z|x,\alpha)$ of hidden variables $z$ (such as damage states and hazard occurrences) given observations $x$ (DPM measurements) and parameters $\alpha$. This is done by introducing a simpler distribution $q(z_{1:m}|\lambda)$ with variational parameters $\lambda$, which we optimize to closely match the true posterior. The optimization objective is to minimize the Kullback-Leibler (KL) divergence between these distributions:

\begin{equation*}
    KL(p||q) = \mathbb{E}_{q}[\text{log}\frac{q(z)}{p(z|x)}]
\end{equation*}

We may not be able to minimize the KL divergence directly, but we can minimize a function equivalent to it up to a constant, known as the evidence lower bound (ELBO). Recall Jensen's inequality as applied to probability distributions. When the function f is concave:

\begin{equation*}
    f(\mathbb{E}[X]) \geq \mathbb{E}[f(X)].
\end{equation*}

By applying Jensen's inequality to the log probability of the observations, we can obtain the ELBO \cite{jordan1999introduction}:

\begin{equation} 
\begin{aligned}
\log p(x) & = \log \int_{z} p(x,z) \\
 & = \log \int_{z} p(x,z) \frac{q(z)}{q(z)}\\
 & \geq \mathbb{E}_{q}\log p(x,z) - \mathbb{E}_{q}\log q(z)
\end{aligned}
\end{equation}

We select a family of variational distributions such that the expectations are computable. Then, we maximize the ELBO to find the parameters that provide as tight a bound as possible on the marginal probability of $x$. Observe that $\text{log}p(x)$ does not depend on $q$. Thus, as a function of the variational distribution, minimizing the KL divergence is equivalent to maximizing the ELBO \cite{hoffman2013stochastic}.

This approach is particularly valuable for our problem because it allows us to efficiently handle the complex interactions between multiple hazards and damage states while maintaining computational feasibility at regional scales. The variational framework also provides a natural way to incorporate our prior knowledge about hazard relationships and building vulnerability.

\subsection{Causal graph-based Bayesian network}
Our method generalizes the formulation and inference of different types of random variables and their complex dependent relationships, such as dependencies and mutual exclusiveness among hazards, dependencies among intermediate hazards, impacts, and sensing observations. In QVCBI, we apply causal Bayesian inference to specific seismological cases. To model the graphical network, we represent the disaster chain of the relationships among initial disasters, secondary hazards and impacts, and sensing observations in three causally-related layers. The initial disasters, such as earthquakes, lead to subsequent hazards and impacts. These secondary hazards and impacts result from the initial disasters. For example, an initial earthquake can cause building damage, landslides, liquefaction, and environmental changes. Due to these hazards and impacts, sensing observations, such as satellite data, are captured by sensing systems.

As depicted in Figure \ref{BN_all}(b), we define our disaster chain as three causally-linked layers consisting of the initial disaster (earthquakes), secondary hazards (building damage, landslide, and liquefaction), and changes in sensing observations to abstract the causal dependencies. We formulate this causal graph-based Bayesian network to represent the statistical causal relationships among the initial disasters, intermediate hazards, prior physical models, and sensing observations. Then, we apply the Bayesian network to infer the probability distribution of building damage and ground failures caused by earthquakes. Our goal is to formulate the causal graph as a Bayesian network using the aforementioned causal graph. By applying the Bayesian network, we can infer the posterior distribution of building damage and earthquake-induced ground failures based on sensing observations. To formulate the causal graph as a Bayesian network, we model vertices as random variables and edges as statistical conditional dependencies that quantify the causal relationships between variables. To improve estimation accuracy and reduce uncertainty, we introduce prior physical information and inventory data with causal dependencies into our Bayesian network.

The Bayesian network consists of two types of vertices: feature vertices and weight vertices. As shown in Figure \ref{BN_all}(a), feature vertices are divided into two classes: unobserved variables $X$, which include hazards and impacts, and observed variables $Y$, which encompass sensing data and prior physical information. These feature vertices vary with location. Without loss of generality, we can assume the hazard/impact variables $X$ follow categorical distributions. For example, we can denote $X_{Building Damage}^{l}$ as an (M+1)-class distribution referring to whether there is building damage with the severity of $m \in {0, \cdots, M}$ occurring at location $l$. Weight vertices $W$ contain parameters that quantify the causal relationships among feature vertices. Three types of weights that depict different causal relations are included in our model: (1) unknown weights describing causal dependencies among unobserved random variables; (2) unknown weights describing causal dependencies between unobserved random variables and observed variables, and (3) predetermined weights assigned from prior physical knowledge or inventory data, which are regarded as hyperparameters. We assume that the values of these weight vertices are consistent across different locations. Given a map with several locations (N), we provide an example of a seismic event in Figure \ref{BN_all}(d).

\begin{figure*}[t]
    \centering
    \includegraphics[width=1\textwidth]{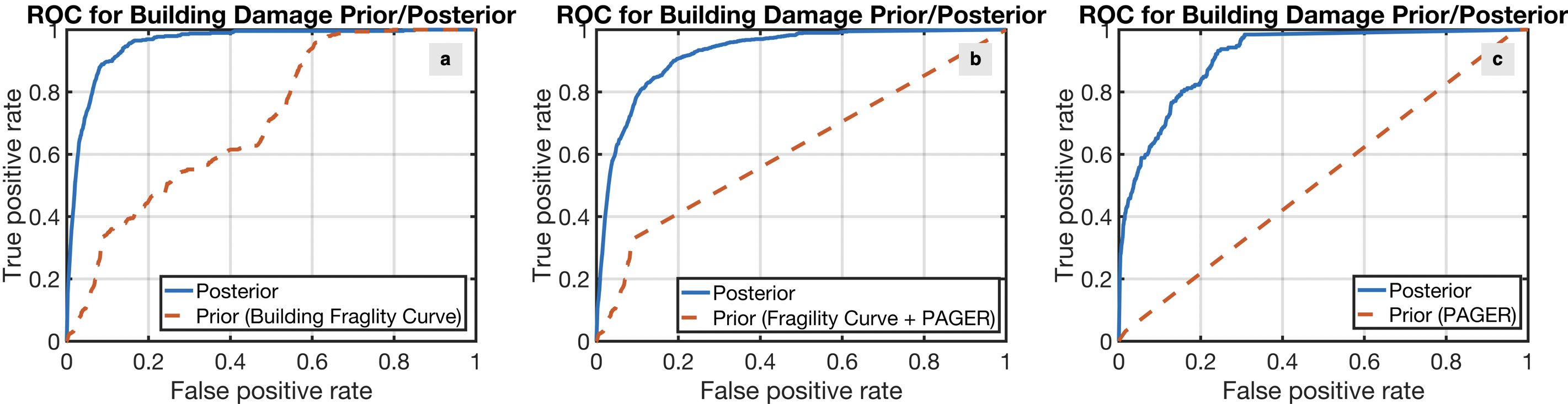}\caption{\textbf{ROC curves of Puerto Rico building damage posterior under different prior models.} Figure (a) displays the ROC curves of the posterior and the HAZUS prior. Figure (b) shows the ROC curve of the posterior and the ROC curve of the combined the HAZUS and PAGER prior model. Figure (c) presents the ROC curves of the posterior model and the prior model from PAGER. }
    \label{PR_bin_ROC}
\end{figure*}

With the modeled Bayesian network, we aim to estimate the probability distributions of unobserved intermediate hazards and impacts. We use $y_{l}$ to refer to a leaf node, $DPM_{l}$, where $l$ represents the location of the $DPM$ data point. Given $y$, we use $\mathcal{P}(y_{l})$ to define the parents of $y_{l}$ at location $l$. We assume the mapping function from parents to $y_{l}$ to be log-normal, as follows:

\begin{equation*}
    \text{log}y|x_{\mathcal{P}(y)} \sim \mathcal N(\sum_{k \in {\mathcal{P}(y)}}w_{ k,y,m_{k}}x_k + w_{\epsilon}\epsilon_{y} + w_{0,y}, {w^2}_{\epsilon_{y}})
\end{equation*}

The hidden damage nodes: building damage(BD), landslide(LS), and liquefaction(LF), are multi-categorical variables $x_{i}$ $\in$ $\{0,1,..., M_{i}\}$, has $M_{i} + 1$ multiple values in total, where $i \in \{BD, LS, LF\}$. For notational simplicity, we define a leak node, with index 0, that is always active ($x_{0} = 1$). It allows its child nodes to be active even if other parent nodes are inactive. For example, even if neither landslide nor liquefaction is present, it is still possible to have building damage due to the ground shaking alone. All nodes are linked by an arbitrary directed graph, where $\mathcal{P}(i)$ are parents of node $i$ (excluding the leak node). The activation probabilities are defined as follows: 

\begin{equation} 
\begin{aligned}
    \text{log}\frac{p(x_{i} = m_{i}|x_{\mathcal{P}(i)},\epsilon_{i})}{p(x_{i} = M_{i}|x_{\mathcal{P}(i)},\epsilon_{i})} = E_{m_{i}}  & \\
  = \sum_{k \in {\mathcal{P}(i)}}w_{k,i, m_{k}}x_k & + w_{\epsilon_i, m_i}\epsilon_{i, m_i} + w_{ 0,i, m_i}
\end{aligned}
\label{Emi}
\end{equation}

\noindent so the probability of $x_{i} = m_{i}$ given the parents of node $i$ is: 

\begin{equation*}
    p(x_i = m_i|x_{P(i)}, \epsilon_i) = \frac{\text{exp}( E_{m_{i}})}{\sum_{m_i} \text{exp}(E_{m_{i}})}
\end{equation*}

\noindent where $w_{  k,y, 0} = w_{  \epsilon_i, 0} = w_{  0,i, 0} = 0, \forall k \in \mathcal{P}(i), \forall i \in \{LS, LF, BD\}$. If all parents are active ($x_{k} = 1; \forall k \in \mathcal{P}(i)$), they active the child node $i$ with probability of

\[ p(x_i = m_i|x_{\mathcal{P}(i)}, \epsilon_i) = \frac{\text{exp}( E_{m_{i}})}{\sum_{m_i} \text{exp}(E_{m_{i}})}\]

\noindent regardless of the states of other parents. If $x_{k} = 0$, parent $k$ has no influence on the state of $x_{i}$.

 \begin{figure*}[t]
    \centering
    \includegraphics[width=1.1\textwidth]{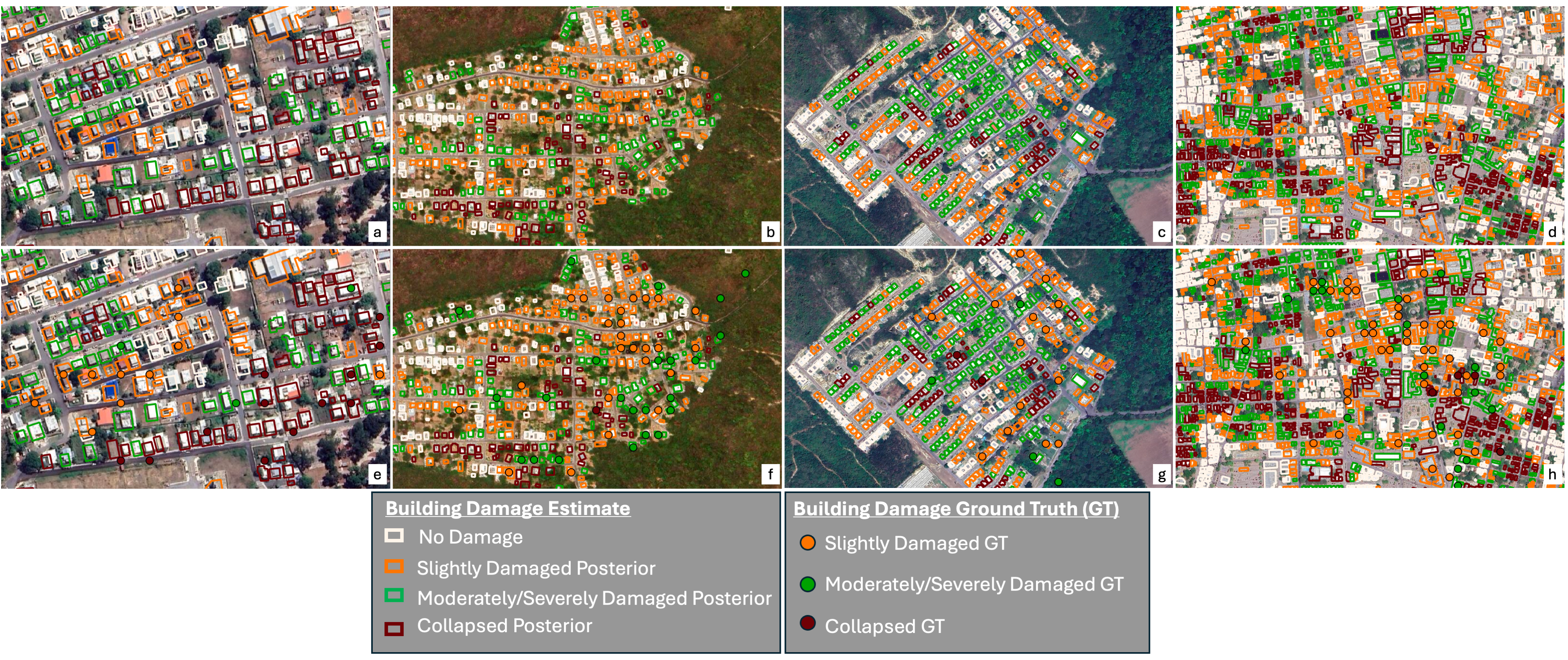}
    \caption{Multi-class building damage assessment results for the 2020 Puerto Rico earthquake. Top row (a-d) shows our posterior estimates, while bottom row (e-h) shows the same estimates overlaid with ground truth data (GT, shown as dots). The comparison across different urban areas demonstrates the agreement between our estimations and actual damage patterns.}
    \label{PR_mul}
\end{figure*}

An XOR node $u$ is also set up to refer to the mutually exclusive states between its parents: Liquefaction ($x_{1}: LF$) and Landslide ($x_{2}: LS$). The parents of $u$ are defined as $\mathcal{P}(u)$. The distribution of $u|x_{\mathcal{P}(u)}$ is a Kronecker delta function, which is defined as:

\begin{equation*}
    u|x_{\mathcal{P}(u)} = \left\{
        \begin{aligned}
        1 & , & \text{if } u = \prod_{k \in \mathcal{P}(u)}x_{k}, \\
        0 & , & \text{otherwise}.
        \end{aligned}
        \right.
\end{equation*}

Due to the difficulty of optimizing the posterior based on the discrete Kronecker data function, it is first transferred into its continuous version, which is a Dirac delta function $\delta(u - \prod_{k \in \mathcal{P}(u)}x_{k})$. Then, the Gaussian distribution is used to approximate the Dirac delta function. The approximate distribution of $u|x_{\mathcal{P}(x)}$ is:

\begin{equation*}
    p(u|x_{\mathcal{P}(u)}) = \frac{1}{\sqrt{2\pi}\sigma}\text{exp}[-\frac{(u - \prod_{k \in \mathcal{P}(u)}x_{k})^{2}}{2\sigma^{2}}]
\end{equation*}

\noindent where $\sigma$ is a small real positive number, $\sigma^{2} \rightarrow 0$.

With the above assumptions and distribution, the Bayesian network in Figure \ref{BN_all}(d) is formulated based on the causal graph that effectively captures the dependencies between different ground failure types, building damage, and remote sensing observations. With the causal Bayesian network constructed, the next step is to infer the posteriors of hazards and impacts. The inference step is introduced in the following subsection.

\subsection{Quadratic variational lower bounds in Causal Bayesian Inference}

In our multi-hazard assessment framework, we need to efficiently estimate building damage and earthquake-induced ground failures across large geographical areas. This presents two key challenges: handling the complex interactions between multiple categorical variables (different damage states and hazard types) and maintaining computational feasibility at scale. To address these challenges, we develop QVCBI using quadratic variational lower bounds, which provide tighter approximations than traditional linear bounds while remaining computationally tractable.

\subsubsection{Variational Distribution Formulation}
For each location $l$, a variational distribution is defined as $q(X^{l})$. We first define an indicator function, where:

\begin{equation*}
\mathbb{I}(x_{i} = m_{i}) = \left\{
        \begin{aligned}
        1 & ,  \text{if} & x_{i} = m_{i}, \\
        0 & , & \text{otherwise.}
        \end{aligned}
        \right.
\end{equation*}

To ensure scalability, we conduct the variational inference on small batches of randomly sampled locations in each iteration. For each location $l$, ${q^{l}}{i, m{i}}$ approximates the posterior probability that node $i$ has value $m_i$:

\begin{equation*}
    {q^{l}}_{i, m_{i}} = p(x_{i}^{l} = m_{i}|x_{\mathcal{P}(i^{l})}, \epsilon_{i}^{l})
\end{equation*}

\noindent where $i^{l} \in {LS, LF, BD}$ represents landslides, liquefaction, and building damage respectively, and $x_{i}^{l} \in {0, \cdots, M_{i}}$ denotes their possible states. The variational distribution factorizes over hidden nodes as:

\begin{equation*}
    q(X^{l}) = {\prod_{i}({q^{l}}_{i, m_{i}})}^{\mathbb{I}({x^{l}}_{i} = m_{i})}
\end{equation*}

\subsubsection{Lower Bound Derivation}

At each geo-location $l$, $q_{i, m_{i}}^{l}$ is defined to approximate the posterior probability that node i with value $m_{i}$ is active in location $l$. We fix $q_{i, 0}^{l} = 1$ so the leak node is always on. For any $q(x^{l})$, the marginal log-likelihood of the observed DPM $y^{l}$ can be lower bounded by Jensen's inequality as follows:

\begin{equation} 
\begin{aligned}
\text{log}p(Y,U) & = \sum_{l \in \mathcal{L}}\text{log}\int p(y^l,u^l,X^l,\epsilon^l)d(X^l,\epsilon^{l}) \\
 & \geq \sum_{l \in \mathcal{L}} \int q(X^l,\epsilon^l) \text{log}\frac{p(y^l,u^l,X^l,\epsilon^l)}{q(X^l,\epsilon^l)}d(X^l,\epsilon^l) \\
  & = \sum_{l \in \mathcal{L}}(\int q(X^l,\epsilon^l)\text{log}p(y^l,u^l,X^l,\epsilon^l)d(X^l,\epsilon^l) - \int q(X^l,\epsilon^l)\text{log} q(X^l,\epsilon^l)) \\
 & = \sum_{l \in \mathcal{L}}(\underbrace{\mathbb{E}_{q(X^l,\epsilon^l)}\text{log}p(y^l,u^l,X^l,\epsilon^l)]}_{[1]} - \underbrace{\mathbb{E}_{q(X^l,\epsilon^l)}[\text{log} q(X^l,\epsilon^l)]}_{[2]})
\end{aligned}
\label{logY}
\end{equation}

\noindent We further expand item [1] in Equation \ref{logY} as:

\begin{equation}
\begin{aligned}
 & \mathbb{E}_{ \begin{subarray}{c} X^{l} \sim q(X^{l})\\ \epsilon^{l} \sim \mathcal N(0,1)\end{subarray}}[\text{log}p(y^{l},u^{l},X^{l},\epsilon^{l})] \\
 & = \underbrace{\mathbb{E}_{\begin{subarray}{c} x_{\mathcal{P}(y^l)}^l \sim q(\mathcal{P}(y^l))\\  \epsilon_{y}^l \sim \mathcal N(0,1)\end{subarray}}[\text{log}p(y^l|x_{\mathcal{P}(y^l)}^l,\epsilon_{y}^l)]}_{[3]} \\
 & + \underbrace{\sum_{i,m_i}\mathbb{E}_{\begin{subarray}{c}  x_{i,m_i}^l \sim q(x_{i,m_i}^l)\\ x_{\mathcal{P}(i^{l})} \sim q(x_{\mathcal{P}(i^{l}),m_{\mathcal{P}(i^{l})}})\\  \epsilon_{i,m_i}^l \sim N(0,1) \end{subarray}}[\text{log}p(x_{i,m_i}^l|\epsilon_{i}^l,x_{\mathcal{P}(i)})]]}_{[4]}\\
 & + \underbrace{\mathbb{E}_{x_{\mathcal{P}(u^l), m_{\mathcal{P}(u^l)}}}[\text{log}p(u^l|x_{\mathcal{P}(u^l), m_{\mathcal{P}(u^l)}})]}_{[5]}\\
 & + \underbrace{\sum_{i,m_i} \mathbb{E}_{\epsilon_{i,m_i}^l \sim \mathcal N(0,1)}[\text{log}p(\epsilon_{i,m_i}^{l})] \mathbb{E}_{\epsilon_{y}^{l} \sim \mathcal N(0,1)}[\text{log}p(\epsilon_{y}^{l})]}_{C_1,fixed}\\
\end{aligned}
\label{expand}
\end{equation}

The item calculation is the most formidable task when computing the variational lower bound [4]. As for item [4], we consider two scenarios - when $x_{i}$ is a leaf node and a non-leaf node. First, we describe the conditional distribution of BD, LS, LF as follows:

\begin{equation}\label{eq3}
p(x_{i}^l=m_{i}|x_{\mathcal{P}(i), m_{\mathcal{P}(i)}}^{l}, \epsilon_{i}^l) = \frac{E_{m_{i}}}{\sum_{m_{i}} E_{m_{i}}} = q_{i,m_{i}}
\end{equation}
\noindent where $E_{m_{i}}$ is defined is in Equation \ref{Emi}. 

The the logarithm of Equation \ref{eq3} can be formulated as: 

\begin{equation}\label{eq4}
\begin{aligned}
& \text{log}p(x_{i, m_{i}}|x_{\mathcal{P}(i), m_{\mathcal{P}(i)}}, \epsilon_{i}^l) \\
& = \text{log}[\frac{\text{exp}(\sum_{k}w_{k, i, m_{i}}x_{k}^l + w_{\epsilon_{i}, m_{i}}\epsilon_{i, m_{i}} + w_{0,i, m_{i}})}{\sum_{m_{i}} \text{exp}(\sum_{k}w_{k, i, m_{i}}x_{k}^l + w_{\epsilon_{i}, m_{i}}\epsilon_{i, m_{i}} + w_{0,i, m_{i}})}] \\
& = \sum_{k \in {\mathcal{P}(i)}}w_{k, i, m_{i}}x_{k}^l + w_{\epsilon_{i}, m_{i}}\epsilon_{i, m_{i}} + w_{0,i, m_{i}} \\
& - \text{log}(\sum_{m_{i}} \text{exp}(\sum_{k \in {\mathcal{P}(i)}}w_{k, i, m_{i}}x_{k}^l + w_{\epsilon_{i}, m_{i}}\epsilon_{i, m_{i}} + w_{0,i, m_{i}}))
\end{aligned}
\end{equation}

\subsubsection{Novel Quadratic Bounding Approach}

\begin{table}[t]
 \centering
 \caption{AUC results for the  building damage estimation under different scenarios in the 2021 Haiti earthquake case.}
 \begin{tabular}{l c}
\toprule
Model  & $AUC$  \\
\toprule
\textbf{Our posterior with the HAZUS prior} & \textbf{0.9412}\\
Our posterior with the PAGER prior & 0.9111\\
Our posterior with the combined prior & 0.9215\\
HAZUS prior & 0.8300\\
PAGER prior & 0.5081\\
Combined prior & 0.8256\\
VBCI \cite{xu2022seismic} & 0.9071\\
Ensemble \cite{rao2023earthquake} & 0.8912\\
 \bottomrule
 \end{tabular}
 \label{Haiti_BD}
 \end{table}

The distribution of the log-sum-exp term in Equation \ref{eq4} is intractable as it contains both discrete and continuous variables. We develop a novel two-stage bounding approach to obtain a tight approximation:

First, we define:

\begin{equation*}
    z_{m_{i}} = \sum_{k \in {\mathcal{P}(i^l)}}w_{k, i, m_{i}}x_{k}^l + w_{\epsilon_{i}, m_{i}}\epsilon_{i^l, m_{i}} + w_{0,i, m_{i}}
\end{equation*}

\noindent and 

\begin{equation*}
  f(z) = \text{log}(\sum_{m_{i}}\text{exp}(z_{m_{i}}))
\end{equation*}

To obtain a better approximation, we adopt a new bound for the log-sum of the exponential obtained by two simple stages: First, the sum of the exponential is upper bounded by a product of sigmoids. Then, the standard quadratic bound on $\text{log}(1+e^{x})$ is used to obtain the final bound. Since the fact that for any $x \in \mathbb{R}$:

  \[
    \prod_{m_{i} = 0}^{M_{i}}(1 + \text{exp}(z_{m_{i}} - \alpha))
    \geq \sum_{m_{i} = 0}^{M_{i}}\text{exp}(z_{m_{i}} - \alpha)
    = e^{- \alpha}\sum_{m_{i} = 0}^{M_{i}}e^{z_{m_{i}}}
    \]
    
\noindent we have:

\begin{equation} \label{eq5}
    \text{log}(\sum_{m_{i} = 0}^{M_{i}}e^{z_{m_{i}}})
    \leq \alpha + \sum_{m_{i} = 0}^{M_{i}}\text{log}(1 + e^{z_{m_{i}} - \alpha})    
\end{equation}

The key property of this bound is that its asymptotes are parallel in most directions. More exactly, by applying the bound to $ax$ where $a \rightarrow \infty$, the difference between the right and the left part of the equation tends to a constant if there exists at least one $x_{k}$ positive and $x_{k} \neq x_{k'}$ for all $k \neq k'$. The above will be relevant when we want to compute the expectation of this function, i.e., item [4], assuming that $\textit{x}$ is a multivariate random variable with high variance \cite{bouchard2008efficient}.
With the above assumptions, we can apply the standard quadratic bound for $\text{log}(1+\text{exp})$ \cite{jaakkola1997variational}:

\begin{equation*}
       \text{log}(1 + e^{z})  \leq \lambda(\xi)(z^{2} - \xi^{2}) + \frac{z - \xi}{2} + \text{log}(1 + e^{\xi}) 
\end{equation*}

\noindent for all $\xi \in \mathbb{R}$, where $\lambda(\xi) = \frac{1}{2\xi}[\frac{1}{1+e^{-\xi}} - \frac{1}{2}]$. 
Apply it inside Equation \ref{eq5}. For any $\textbf{x} \in \mathbb{R}^{M_{i + 1}}$, any $\alpha \in \mathbb{R}^{M_{i + 1}}$, and any $\xi \in [0,\infty)^{K}$, we have:

\begin{equation} \label{bound}
\begin{aligned}
    \text{log}(\sum_{m_{i} = 0}^{M_{i}}e^{z_{m_{i}}})
    & \leq \alpha + \sum_{m_{i} = 0}^{M_{i}}[\lambda(\xi_{m_{i}})((z_{m_{i}} - \alpha)^{2} - \xi_{m_{i}}^{2}) \\
    & + \frac{z_{m_{i}} - \alpha - \xi_{m_{i}}}{2} + \text{log}(1 + e^{\xi_{m_{i}}})]
\end{aligned}
\end{equation}

\noindent The minimization of the upper bound with respect to $\xi$ gives:

\begin{equation*}
\begin{aligned}
    [-\frac{1}{2\xi_{m_{i}}^{2}}(\frac{1}{1+e^{-\xi_{m_{i}}}} - \frac{1}{2}) + \frac{e^{-\xi_{m_{i}}}}{2\xi_{m_{i}}(1+e^{\xi_{m_{i}}})^2}][(z_{m_{i}} - \alpha_{i})^{2} & \\ - \xi_{m_{i}}^{2}] 
    - 2\xi_{m_{i}}\lambda(\xi_{m_{i}}) - \frac{1}{2} + \frac{e^{\xi_{m_{i}}}}{1 + e^{\xi_{m_{i}}}} = 0
\end{aligned}
\end{equation*}

\noindent for $m_{i} = 1,2,\cdots,M_{i}$, $i \in \{LS, LF, BD\}$. The minimization with respect to $\alpha_{i}$ gives:

\begin{equation}
    \hat{\alpha}_{i} = \frac{4\sum_{m_{i} = 0}^{M_{i}}z_{m_{i}}\lambda(\xi_{m_{i}}) - (1 - M_{i})}{4\sum_{m_{i} = 0}^{M_{i}}\lambda(\xi_{m_{i}})}
\label{alpha_hat}
\end{equation}

\noindent Substitute Equation \ref{alpha_hat} into Equation \ref{bound}, we can get the lower bound on $\text{log}(\sum_{m_i = 0}^{M_{i}}\exp(z_{m_{i}}))$. The remaining derivation is shown in Appendix \ref{Appendix_lower_bound}.

\subsubsection{Final Lower Bound}
By optimizing the variational parameters $\alpha$ and $\xi$, we obtain the final lower bound for the log-likelihood across all locations given a map with a set of locations $L$:

\begin{equation}
\begin{aligned}
    & \mathcal{L}(\textbf{q,w})  = \text{log}P(Y,U) = \sum_{l \in \mathcal{L}}\text{log}P(y^{l},u^{l}) \\
    & \geq \sum_{l \in \mathcal{L}} \{-\text{ln}y^{l} - \text{ln}|w_{\epsilon_{y}}| - \frac{1}{2}\text{ln}2\pi + \sum_{i, m_{i}}m_{i}q_{i, m_{i}}[ \mathbb{E}(z_{m_{i}}) + \mathbb{E}(\hat{\alpha}^{2})\sum_{m_{i} = 0}^{M_{i}}\lambda(\xi_{m_{i}})\\
    & -\sum_{m_{i} = 0}^{M_{i}}\text{log}(1+e^{\xi_{m_{i}}}) - \sum_{m_{i} = 0}^{M_{i}}\lambda(\xi_{m_{i}})(\mathbb{E}(z_{m_{i}}^{2})-\xi_{m_{i}}^{2})]\\
    & -\sum_{i, m_{i}}\sum_{m_{i} = 0}^{M_{i}}\frac{1}{2}m_{i}q_{i, m_{i}}(\mathbb{E}(z_{m_{i}}) - \xi_{m_{i}}) -\frac{1}{2}\text{log}2\pi \sigma^2 \\
    & - \frac{\prod_{k\in \mathcal{P}(u^l)}\sum_{m_{k} = 0}^{M_{k}}m_{k}^2 q^l_{k,m_{k}}}{2\sigma^2}  - \sum_{i,m_{i}}q_{i,m_i}\text{log}q^l_{i,m_i}\\
    & - \frac{\sum_{\begin{subarray}{c} i,j \in \mathcal{P}(y^l)\\ m_{i}, m_{j}\\ i \neq j\end{subarray}}w_{i,y,m_{i}}w_{j,y,m_{j}}(m_{i}q^l_{i,m_{i}})(m_{j} q^l_{j,m_{j}})}{w_{\epsilon_{y}}^2} \\ 
    & - \frac{(\text{ln}y^{l})^2 + w_{0,y}^2 + w_{\epsilon_{y}}^2 + \sum_{k \in \mathcal{P}(y^{l}), m_{k}}w_{k,y,m_{k}}^2 m_{k}^2 q_{k,m_{k}}}{2w_{\epsilon_{y}}^2} \\
    & - \frac{(w_{0,y}-\text{ln}y^l)\sum_{k \in \mathcal{P}(y^l), m_{k}}w_{k,y,m_{k}} m_{k}q_{k, m_{k}}- w_{0,y}\text{ln}y^l}{w_{\epsilon_{y}}^2} \}
\label{ELBO}
\end{aligned}
\end{equation}

\noindent where $\lambda_{\xi}$, $z_{m_{i}}$, $\hat{\alpha}$ are defined as above. With the tight lower bound of log-likelihood of DPM observations ,we can further maximize the lower bound to find the optimal posteriors of unobserved variables, i.e., LS, LF and BD. The details of the variational lower bound derivation in Equation \ref{ELBO} is shown in Appendix \ref{Appendix_lower_bound}.

\subsection{L-1 regularization for trade-off between DPM and prior geospatial models}
A key challenge in QVCBI is balancing the influence of DPMs (Damage Proxy Maps) and prior geospatial models, as each data source has its own limitations. Prior geospatial models may be less accurate in areas with sparse features or unusual seismic patterns, while DPM data can be noisy in regions with complex topography. To address this, we introduce L-1 regularization to control the relative influence of each data source on our final estimates.

The regularization objective is formulated as:

\begin{equation*}
    \min_{\begin{subarray}{c}w\\\end{subarray}}\mathcal{L}_{1} =  \min_{\begin{subarray}{c}w\\\end{subarray}} \lambda_{1} \sum_{\begin{subarray}{c} i \in \mathcal{P}(y)\\ i \neq \{x_{0}, \epsilon_{y}\}\end{subarray}}|w_{iy}| + \lambda_{2} \sum_{\begin{subarray}{c}i \in \{LS,LF\}\\\end{subarray}}|w_{\alpha i}|
\end{equation*}

where $\lambda_{1}$ and $\lambda_{2}$ are regularization parameters that control the influence of DPMs and geospatial models respectively. Increasing either parameter constrains the influence of its corresponding data source. The final optimization objective combines this regularization term with our variational lower bound:

\begin{equation*}
    \max_{\begin{subarray}{c}\text{q,w}\\\end{subarray}}\mathcal{L}(\text{q,w})- \lambda_{1}\sum_{\begin{subarray}{c} i \in \mathcal{P}(y)\\ i \neq \{x_{0}, \epsilon_{y}\}\end{subarray}}|w_{iy}| - \lambda_{2} \sum_{\begin{subarray}{c}i \in \{LS,LF\}\\\end{subarray}}|w_{\alpha i}|
\end{equation*}

The optimal $q_{i}^{l}$ and causal coefficients $w$ should be optimized to maximize the variational lower bound of likelihood of DPM while minimizing the $l_{1}$ norm of causal coefficients between the probabilistic graph and DPMs/prior geospatial models.

\begin{table*}
\caption{The evaluation of QVCBI performance when estimating the multi-categorical building damage levels.}
\centering
    \begin{tabular}{l|c|ccc}
        \toprule
         Earthquake Cases & Class & $AUC_{Post}$ & $AUC_{HAZUS}$ & Cross Entropy\\
         \midrule
        & Slight damage & \textbf{0.9413} & 0.8299 & 0.0559 \\
        The 2021 Haiti Earthquakes & Moderate damage & \textbf{0.9498}  & 0.8218  & 0.0619 \\
        & Collapse & \textbf{0.9505 }& 0.8754 & 0.0373\\
        \hline
          & Slight Damage & \textbf{0.9587} & 0.7050 & 0.0211\\
         The 2020 Puerto Rico Earthquakes & Moderate damage & \textbf{0.9596} & 0.7213 & 0.0366 \\
          & Collapse & \textbf{0.9544} & 0.7561 & 0.0190\\
         \hline
        & Slight damage & \textbf{0.9540} & 0.9159 & 0.0320 \\
        The 2020 Zagreb Earthquakes & Moderate damage & \textbf{0.9580} & 0.9242 & 0.0056 \\
        & Collapse & \textbf{0.9314} & 0.9062 & 0.0022 \\
        \bottomrule
    \end{tabular}
    \label{multi_BD}
\end{table*}

\subsection{Expectation-Maximization(EM) algorithm}
While our variational bound provides a theoretical framework for estimating the marginal likelihood of observations, we need an efficient algorithm to optimize both the posterior distributions $q_{i}^{l}$ and the causal coefficients $w$. We develop an Expectation-Maximization (EM) approach that alternates between updating local posterior estimates and global model parameters.
For each iteration, we randomly sample a mini-batch of locations from the map. In the Expectation step, we update the local posteriors for building damage, landslide, and liquefaction by maximizing the variational lower bound. In the Maximization step, we update the global causal coefficients using stochastic gradient updates

The optimal posterior follows:

\begin{equation}\label{EM}
   q_{k, m_{k}}^{l} = \text{exp}(\mathcal{T}(q_{\mathcal{P}(k^{l})},q_{\mathcal{S}(k^{l},\mathcal{C}(k^{l}))},q_{\mathcal{C}(k_{l})}, y^{l}, u_{l}))
\end{equation}

\noindent where $\mathcal{C}(i)$ represents child nodes of $i$, $\mathcal{S}(i,k)$ represents spouse nodes of $i$ at child node $k$ (other parents of the child node), and $\mathcal{T}(\cdot)$ is defined in Appendix \ref{appendix_EM}.

For the weight updates in the Maximization step, we use stochastic gradient updates with point estimates due to the non-conjugate noisy-OR likelihood:

\begin{equation}\label{2}
    w^{(t+1)} = w^{(t)} + \rho \mathcal{A}\nabla L^{(t)}(w)
\end{equation}

\noindent where $\rho$ controls the learning rate and $\mathcal{A}$ is a pre-conditioner \cite{paisley2011discrete}. Detailed derivations of the partial derivatives are provided in Appendix \ref{appendix_EM}.

By alternating between these update steps using Equations \ref{EM} and \ref{2}, we maximize the variational lower bound to approximate the marginal likelihood of the sensing observations. The algorithm converges to optimal combinations of hazard/impact posteriors and causal dependency weights, providing our final estimates of multi-class building damage and secondary hazards.

\begin{figure}[t]
    \centering
    \includegraphics[width=0.48\textwidth]{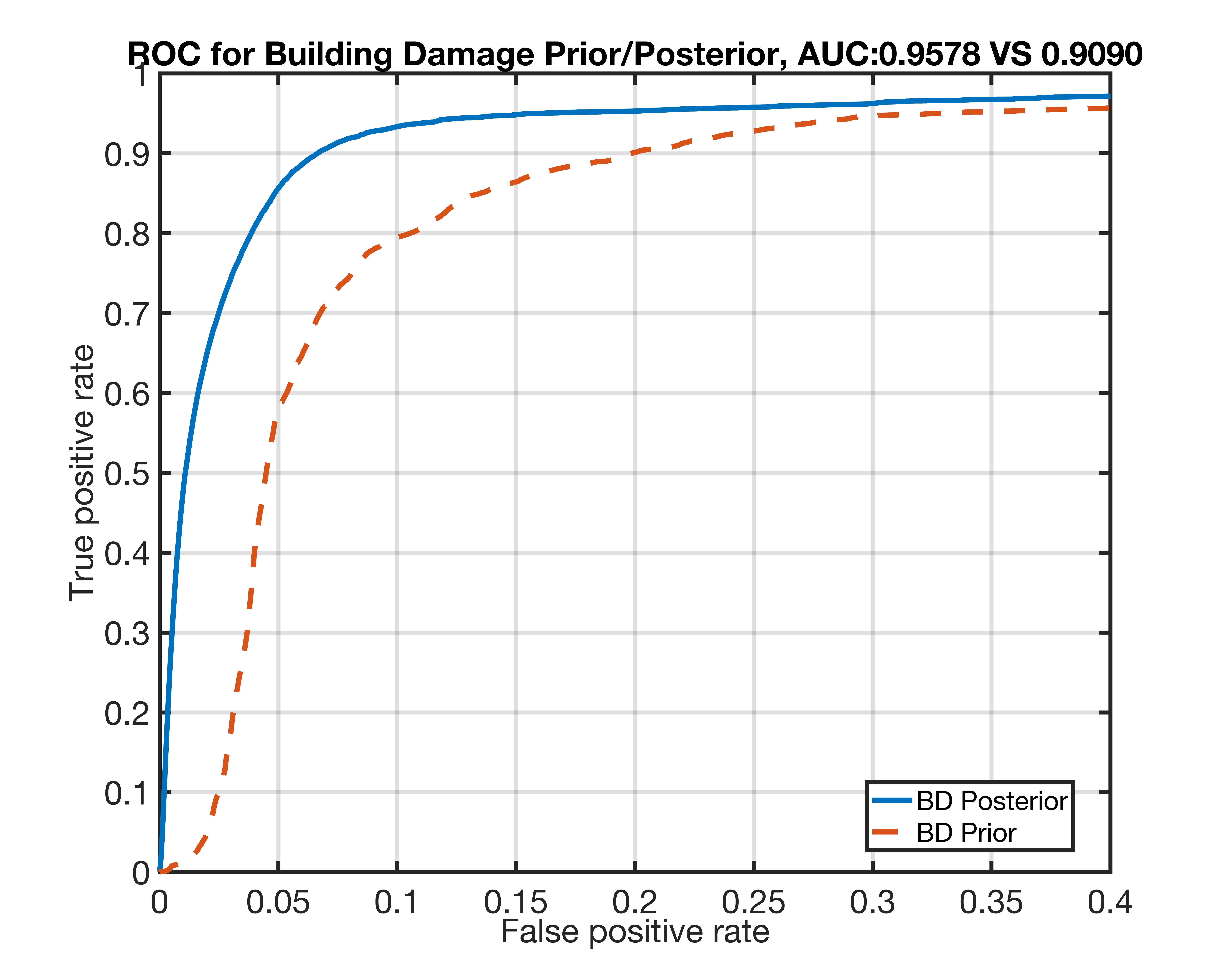}
    \caption{ROC curve of the 2020 Zagreb, Croatia earthquake building damage posterior and the HAZUS prior for building damage.}
    \label{Zagreb_ROC}
\end{figure}

\subsection{Pruning strategy to improve computational efficiency}
To improve computational efficiency and handle the large-scale nature of earthquake damage assessment, we implement a pruning strategy based on building footprint (BF) information. This strategy operates on the principle that areas without buildings cannot experience building damage, allowing us to reduce the computational space by focusing only on locations where buildings are present. The pruning process begins with initial screening of pixels based on BF data, where pixels without building footprints are assigned zero probability of building damage. These pixels are then removed from the active computation set to reduce memory requirements and processing time. After the damage assessment is complete, these pruned pixels are integrated back into the final damage map with zero damage probability assignments.

However, this approach presents challenges when BF data is incomplete or inaccurate, particularly in regions with limited infrastructure documentation. Building footprint data quality varies significantly across different regions and can be particularly problematic in remote or mountainous areas where building documentation may be sparse, rapidly developing regions where building footprint data may be outdated, and areas where informal settlements may not be officially recorded. For example, in the 2021 Haiti earthquake case, Figure \ref{inaccurateBF} illustrates the significant BF data gaps in the mountainous regions in Haiti, where satellite imagery confirms building presence despite missing BF information. Among the 6,272 pixels with reported damaged buildings in our dataset, 2,957 pixels lack BF information, which would theoretically limit traditional assessment methods to a maximum accuracy of 52.85\%.

\begin{figure*}[t]
    \centering
    \includegraphics[width=1\textwidth]{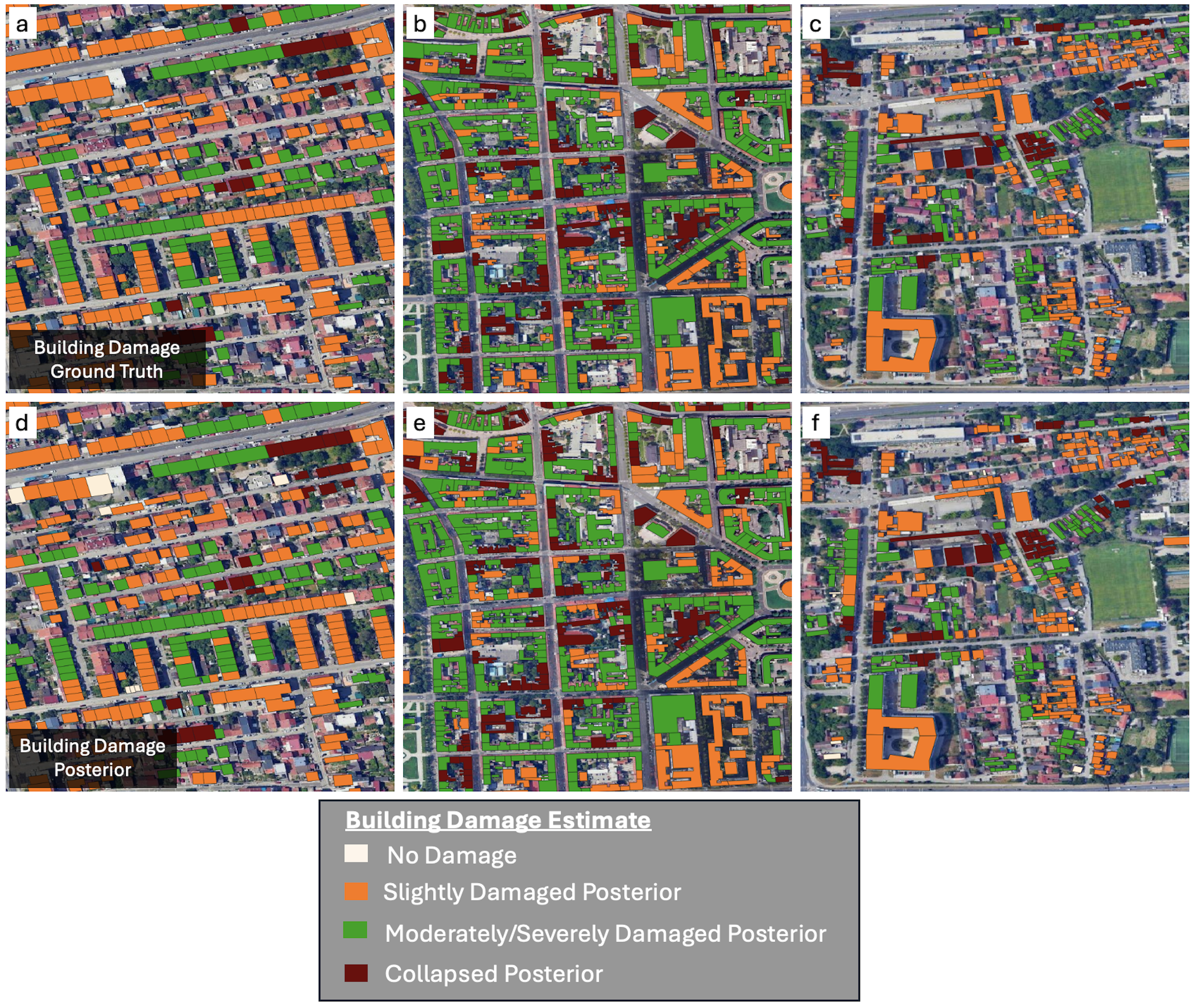}
    \caption{\textbf{Multi-class building damage assessment results for the 2020 Zagreb earthquake.} Top row (a-c) shows ground truth building damage classifications across different urban settings; bottom row (d-f) shows the building damage posterior estimation for the corresponding areas. The comparison demonstrates the accuracy of the model across diverse urban layouts, from regular residential blocks (a,d) to dense city centers (b,e) and mixed urban-suburban areas (c,f).}
    \label{Zagreb_vis}
\end{figure*}

To address these limitations, QVCBI incorporates compensation mechanisms. We utilize prior fragility functions that can suggest building presence even in areas with missing BF data. Additionally, our proposed quadratic variational causal Bayesian inference framework enables itself to overcome initial BF data limitations through evidence from other data sources. As demonstrated in the Haiti case, QVCBI effectively overcomes these data limitations through the combination of prior fragility functions and the quadratic variational causal Bayesian inference approach. The model demonstrates robust performance by adaptively incorporating evidence from multiple data sources, ultimately providing reliable damage assessments even in areas with incomplete BF coverage. This pruning strategy significantly reduces computational overhead while maintaining assessment accuracy in well-documented areas. In regions with incomplete BF data, our compensation mechanisms help maintain reasonable damage estimates despite the data limitations.

\section{Results}

\subsection{Evaluation metrics: AUC and ROC curves}
We evaluate our multi-class building damage estimation method using Receiver Operating Characteristics (ROC) curves and the Area Under the ROC Curve (AUC) metrics \cite{fawcett2006introduction}. These metrics are particularly suitable for assessing the ability of our model to discriminate between different damage classes while accounting for various decision thresholds.
The ROC curve plots the True Positive Rate (TPR, correctly identified damage cases) against the False Positive Rate (FPR, incorrectly classified non-damage cases) across different classification thresholds. This visualization effectively captures the trade-off between sensitivity and specificity in our damage classifications. The AUC, which ranges from 0 to 1, quantifies the overall discriminative ability of the model - a higher AUC indicates better separation between damage classes, with values closer to 1 representing superior performance.

\begin{figure*}[t]
    \centering    \includegraphics[width=1.05\textwidth]{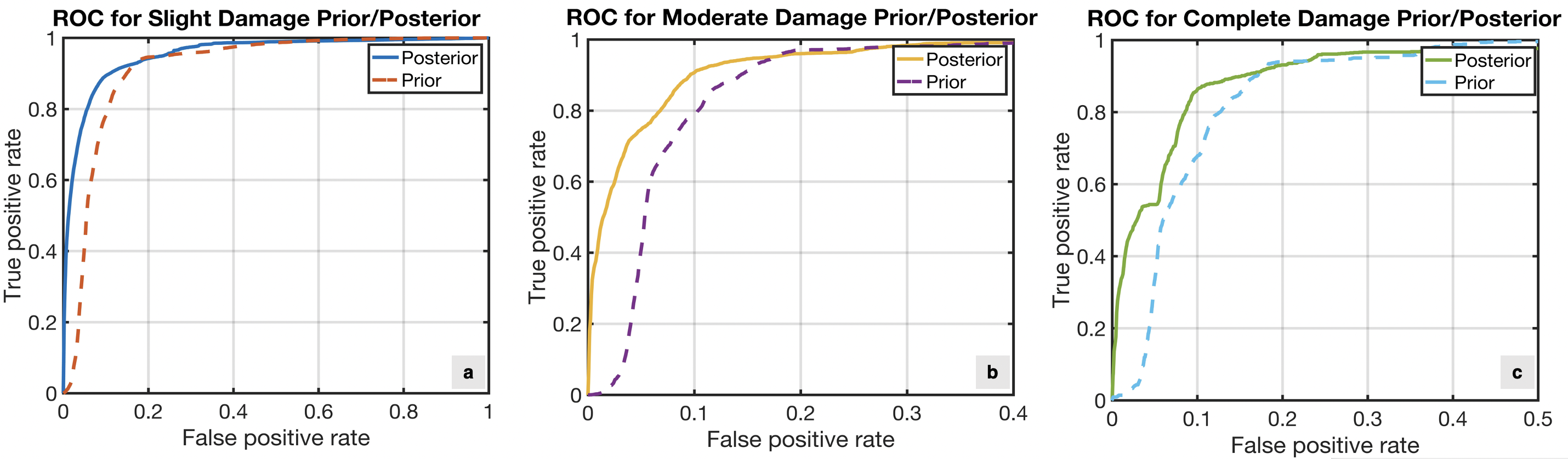}
    \caption{\textbf{ROC curves of the 2020 Zagreb earthquake building damage posterior and prior fragility function.} Figure (a) shows the ROC curve of the posterior and prior model for the slight damage; (b) displays the ROC curve of the posterior and prior model for the moderate damage; and (c) presents ROC curve of the posterior and prior model for the complete damage/ collapse.}
\label{Zagreb_mul_roc}
\end{figure*}

\subsection{Cases analysis and results visualization}
We evaluate the performance of QVCBI using different prior models to demonstrate its robustness and to select the optimal prior model for different scenarios. While QVCBI enables multi-class damage assessment, we first conduct binary classification analysis (damaged vs. undamaged) for two key reasons: to enable direct comparison with existing state-of-the-art methods, which predominantly focus on binary classification, and to validate the fundamental effectiveness of QVCBI before extending to the more complex multi-class scenario. We compare three different prior models: HAZUS prior (based on HAZUS fragility curves), PAGER prior (derived from PAGER system estimates), and the combined prior (linear combination of HAZUS and PAGER probabilities). Building on these binary classification insights, we extend our analysis to the full multi-class damage assessment, categorizing buildings into three stages of building damage: (1) buildings with slight damage; (2) buildings with moderate damage; and (3) buildings with collapse.

\subsubsection{The 2021 Haiti earthquake}

   \begin{figure*}[t]
    \centering
\includegraphics[width=1\textwidth]{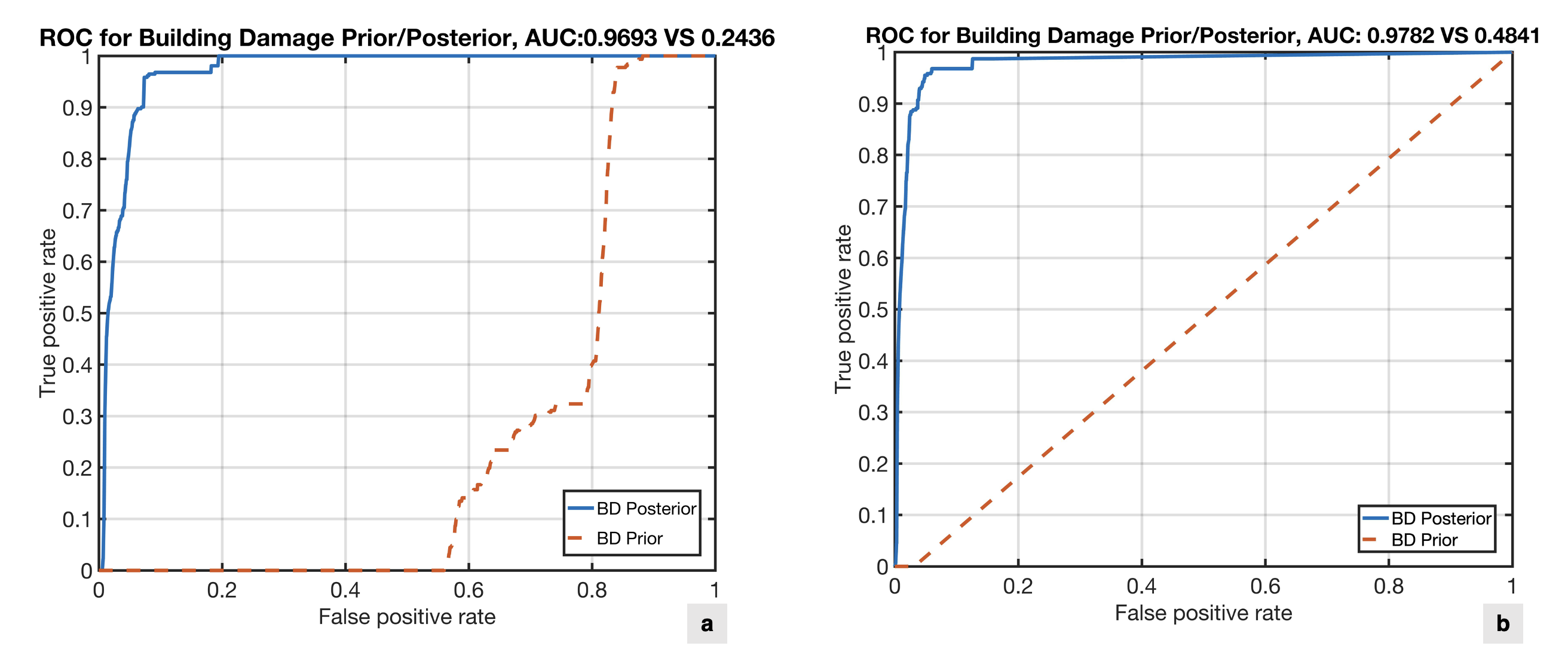}
    \caption{\textbf{ROC curves of the 2016 Italy earthquake building damage posterior and prior fragility function.} Figure (a) shows the ROC curve of the posterior model generated by using the HAZUS prior and the ROC curve of the HAZUS prior. Figure (b) displays the PAGER prior and the ROC curve of our posterior model.}
    \label{Italy_ROC}
\end{figure*}

For the binary case evaluation, as shown in Table \ref{Haiti_BD}, our method demonstrates superior performance across different prior model configurations. When using the HAZUS prior, QVCBI achieves the highest AUC of 0.9412, significantly improving the baseline performance of the HAZUS prior (AUC: 0.8300) by 13.40\%. While the PAGER system alone shows modest performance (AUC: 0.5081), QVCBI substantially enhances its predictive capability (AUC: 0.9111). The combined prior model yields an intermediate improvement (AUC: 0.9215). Notably, our approach outperforms recent advanced methods including VBCI \cite{xu2022seismic} (AUC: 0.9071) and ensemble learning approaches \cite{rao2023earthquake} (AUC: 0.8912). These results demonstrate that QVCBI enhances estimations from various prior models and exceeds the performance of existing state-of-the-art binary classification methods.

For multi-class damage assessment in the Haiti earthquake, Table \ref{multi_BD} shows QVCBI exhibits robust performance across all damage categories. For slight damage, our model achieves an AUC of 0.9413, improving upon the HAZUS prior (0.8299) by 13.42\%. The improvement becomes even more pronounced for moderate damage, where our model reaches an AUC of 0.9498 compared to HAZUS (0.8218), representing a 14.45\% increase. In collapse detection, QVCBI achieves an AUC of 0.9505, surpassing the HAZUS prior (0.8754) by 8.58\%. The consistently low cross-entropy values (0.0559, 0.0619, and 0.0373 for slight, moderate, and collapse categories respectively) further validate the model confidence in predictions. These results demonstrate that QVCBI extends beyond binary classification to provide reliable multi-class damage assessments, addressing a critical gap in current approaches while maintaining high accuracy across all damage levels.

\subsubsection{The 2020 Puerto Rico earthquake}

\begin{table*}[t]
\centering
    \caption{The evaluation of QVCBI robustness under the different prior information for building damage, and see which prior model can result in the most accurate posterior.}
    \begin{tabular}{l|ccc}
        \toprule
        Earthquake Cases & $AUC_{Post}$ & $AUC_{Pr}$ & Cross Entropy\\
        \midrule
         \textbf{Puerto Rico (Using the HAZUS prior)} & \textbf{0.9567} & \textbf{0.7050} & \textbf{0.0302} \\
         Puerto Rico (Using the combined prior model) & 0.9255 & 0.6325 & 0.0491\\
         Puerto Rico (Using the PAGER prior) & 0.9148 & 0.5112 & 0.0613\\

    \midrule
         \textbf{Zagreb (Using the HAZUS prior)} & \textbf{0.9578} & \textbf{0.9090} & \text{0.0397} \\
        Zagreb (Using the combined prior) &  0.9294 & 0.8817 & 0.0413 \\
         Zagreb (Using the PAGER prior) & 0.9220 & 0.8512 & 0.0476\\
    \midrule
         Italy (Using the HAZUS prior) & 0.9693 & 0.2436 & 0.0168\\
         \textbf{Italy (Using the PAGER prior)} & \textbf{0.9782} & \textbf{0.4841} & \textbf{0.0045}\\
        \bottomrule
    \end{tabular}
    \label{differentprior}
\end{table*}

We evaluate model robustness under different prior models for the Puerto Rico earthquake case. Table \ref{differentprior} and Figure \ref{PR_bin_ROC} demonstrate that QVCBI significantly improves upon all tested prior models. Using the HAZUS prior, our model achieves an AUC of 0.9567, representing a 35.7\% improvement over the prior (AUC: 0.7050). The combined prior yields an AUC of 0.9255 (46.32\% improvement from 0.6325), while the PAGER prior results in an AUC of 0.9148 (78.95\% improvement from 0.5112). These results indicate that the HAZUS prior provides the most effective baseline for QVCBI in this case.

For binary classification performance comparison with existing approaches, we evaluate QVCBI against the baseline model from \cite{rao2022earthquake}. Table \ref{conf_PR} reveals superior performance across all metrics, with our model achieving a true positive rate of 0.9987 and a true negative rate of 0.9644, compared to 0.86 and 0.58 respectively for the baseline. These improvements are particularly significant given our substantially larger dataset of 116,548 undamaged and 766 damaged buildings, demonstrating both accuracy and scalability.

The multi-class damage assessment capabilities are visualized in Figure \ref{PR_mul}. The framework successfully distinguishes between four damage states: undamaged buildings (white outlines), slightly damaged structures (orange), moderately to severely damaged buildings (green), and collapsed structures (red). Bottom row comparisons with ground truth data points (shown as dots) provide direct validation of framework accuracy across diverse urban contexts. In dense residential areas (Figures \ref{PR_mul}a,e), the model accurately captures spatial distribution patterns, particularly the concentration of moderate to severe damage in central areas and slight damage in peripheral regions. Less dense regions (Figures \ref{PR_mul}b,f) demonstrate accurate damage classification despite scattered development patterns. Coastal areas (Figures \ref{PR_mul}c,g) show successful identification of varying damage patterns, including clusters of severe damage that may indicate localized seismic amplification effects. Complex urban environments (Figures \ref{PR_mul}d,h) reveal accurate distinction between multiple damage levels despite irregular building arrangements and mixed building types.

Quantitative evaluation of multi-class performance in Table \ref{multi_BD} further validates framework effectiveness. The model achieves consistently high AUC values: 0.9587 for slight damage (improving from HAZUS prior 0.7050), 0.9596 for moderate damage (improving from 0.7213), and 0.9544 for collapse assessment (improving from 0.7561). Low cross-entropy values (0.0211, 0.0366, and 0.0190 for slight, moderate, and collapse categories respectively) confirm estimation reliability. The minimal variation in AUC values across damage levels demonstrates balanced classification capabilities essential for comprehensive damage assessment.

The spatial validation reveals strong agreement between posterior estimates and ground truth across diverse urban contexts. The framework captures both damage presence and severity gradients, demonstrated by smooth transitions between damage levels aligning with ground observations. This comprehensive performance across binary classification, multi-class assessment, and spatial validation demonstrates the practical utility of QVCBI for emergency response applications.

\begin{figure*}[t]
    \centering
    \includegraphics[width=1\textwidth]{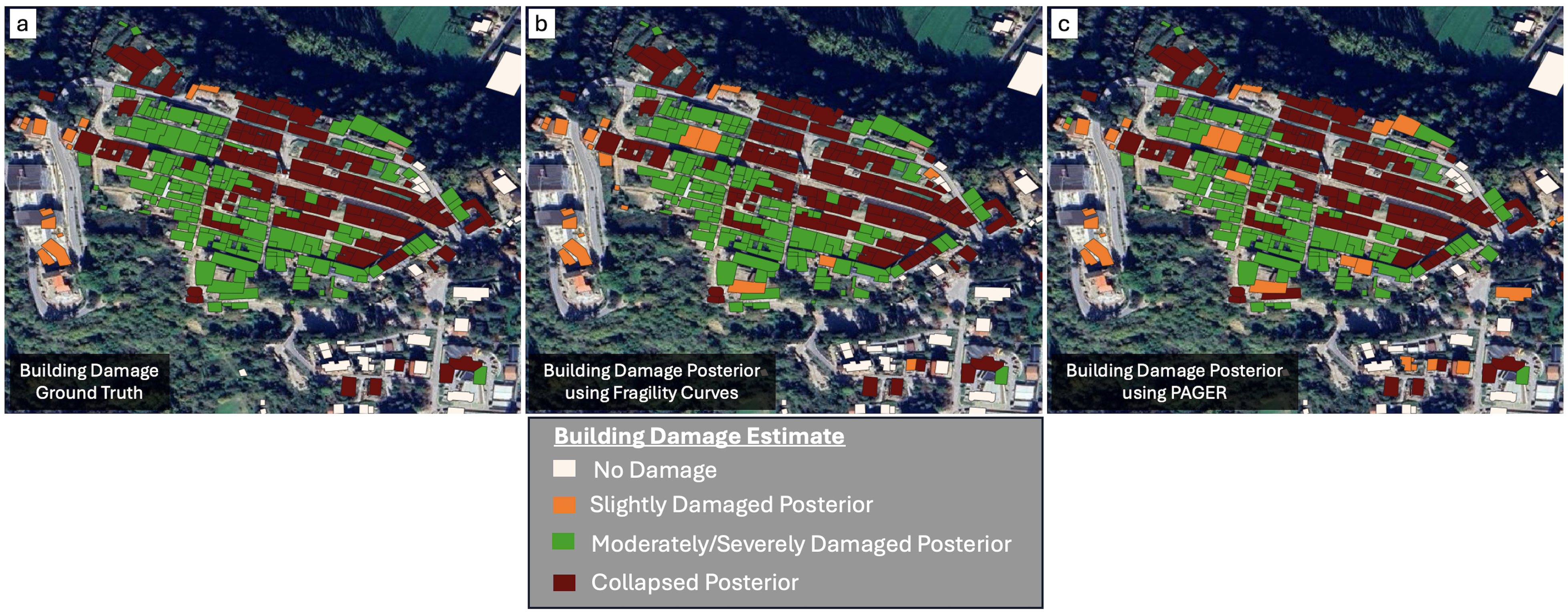}
    \caption{\textbf{Comparison of multi-class building damage assessments for the 2016 Italy earthquake.} (a) Ground truth building damage classification; (b) Building damage posterior using the HAZUS prior; (c) Building damage posterior estimates using the PAGER prior.}
    \label{Italy_mul}
\end{figure*}

\subsubsection{The 2020 Zagreb, Croatia earthquake}

The 2020 Zagreb earthquake provides a valuable case study for evaluating QVCBI in a complex urban environment. For binary classification, we first test model performance using different prior models. Table \ref{differentprior} shows that among the three priors tested (HAZUS, PAGER, and combined), the HAZUS prior yields the best performance. Using the HAZUS prior, our model achieves an AUC of 0.9578, improving upon the prior model (AUC: 0.9090) by 5.37\%. While this improvement appears modest compared to other cases, it is significant given the strong baseline performance of the HAZUS prior in this region. As shown in Figure \ref{Zagreb_ROC}, the improvements are particularly pronounced in the critical low false-positive rate region, indicating better discrimination of damaged buildings while maintaining low false alarm rates.
The confusion matrix comparison with the baseline model \cite{rao2022earthquake} in Table \ref{conf_Zag_1} demonstrates substantial performance improvements. Our model achieves a true positive rate of 0.9899 and a true negative rate of 0.9422, compared to 0.82 and 0.48 respectively for the baseline. These results are particularly noteworthy given our significantly larger dataset of 299,055 undamaged and 18,349 damaged buildings.

For multi-class damage assessment, QVCBI maintains robust performance across all damage categories. As shown in Figure \ref{Zagreb_mul_roc} and Table \ref{multi_BD}, the model achieves consistent improvements over the HAZUS prior across all damage levels: slight damage (AUC: 0.9540 from 0.9159), moderate damage (AUC: 0.9580 from 0.9242), and complete damage/collapse (AUC: 0.9314 from 0.9062). The low cross-entropy values (0.0320, 0.0056, and 0.0022 respectively) further validate estimation reliability. The spatial validation in Figure \ref{Zagreb_vis} demonstrates the effectiveness of QVCBI across diverse urban contexts. In regular residential blocks (a,d), the model accurately captures damage patterns reflecting seismic wave effects on similar building types. Dense urban areas (b,e) show accurate classification despite varying building characteristics and complex spatial arrangements. Mixed urban-suburban regions (c,f) demonstrate reliable damage assessment across different development patterns. This spatial consistency, combined with the strong quantitative metrics, validates the utility of QVCBI for emergency response applications, particularly in complex urban environments where accurate damage assessment is crucial for resource allocation.

\begin{table*}[t]
  \caption{Normalized confusion matrix comparison for binary building damage estimation results produced by QVCBI and the baseline model\cite{rao2022earthquake} in the 2020 Puerto Rico earthquake. TP, FP, FN, TN represent true positive, false positive, false negative, and true negative, respectively.}
 \centering
 \begin{tabular}{|c|c|c|c|c|}
\toprule
Damage Grade &\multicolumn{2}{|c|}{Damaged (Estimated)
} & \multicolumn{2}{|c|}{Undamaged (Estimated)
}\\
\hline
Models & Our Model & Baseline model &  Our Model & Baseline model\\
\hline
Damaged (True) &  \textbf{0.9987 (TP)} & 0.86 (TP)& \textbf{0.0013 (FP)} & 0.14 (FP)  \\
\hline
Undamaged (True) & \textbf{0.0356 (FN)} & 0.42 (FN) & \textbf{0.9644 (TN)}  & 0.58 (FP)\\
\bottomrule
 \end{tabular}
 \label{conf_PR}
 \end{table*}

\subsubsection{The 2016 Italy earthquake}

The 2016 Italy earthquake case study demonstrates the robustness of QVCBI in handling poorly-performing prior models and validates its effectiveness in regions where traditional fragility models struggle. For binary classification, we evaluate our approach using two different prior models: the HAZUS prior \cite{del2017empirical} and the PAGER system. As shown in Table \ref{differentprior}, the results reveal dramatic improvements over both prior models. Using the HAZUS prior, QVCBI achieves an AUC of 0.9693, marking a 298\% improvement over the AUC of the HAZUS prior (0.2436). Figure \ref{Italy_ROC}(a) illustrates this substantial enhancement, showing how our posterior model achieves high true positive rates even at very low false positive rates, in stark contrast to the poor performance of the HAZUS prior. Similarly, with the PAGER prior (Figure \ref{Italy_ROC}(b)), our model maintains excellent performance with an AUC of 0.9782, significantly outperforming the AUC of the PAGER prior (0.4841). These results are particularly noteworthy as they demonstrate the ability of QVCBI to generate reliable damage estimates even when starting with severely underperforming prior models.

The capability of QVCBI extends beyond binary classification to reliable multi-class damage assessment, as demonstrated in Figure \ref{Italy_mul}. The comparison between ground truth (Figure \ref{Italy_mul}a) and damage estimates using different priors (Figures \ref{Italy_mul}b,c) reveals consistently accurate predictions across all damage levels. In dense urban settlements, our model accurately identifies the spatial distribution of collapse patterns (red) and moderate/severe damage areas (green). The framework maintains this high accuracy regardless of the prior model used, with both HAZUS- and PAGER-based estimates showing strong agreement with ground truth.
A key strength demonstrated in the Italy case is the ability of QVCBI to overcome regional variations in building vulnerability that often challenge traditional fragility models. The visual results illustrate accurate damage pattern reconstruction across diverse building types and urban layouts characteristic of Italian cities. This robustness to regional architectural variations, combined with the ability to handle poorly-performing priors, makes QVCBI particularly valuable for rapid damage assessment in regions where existing fragility models may be poorly calibrated to local building stock.
The consistent performance across both binary and multi-class assessments, despite starting with significantly underperforming priors, validates the potential of QVCBI for reliable deployment in international contexts where prior models may be limited or unsuitable. This capability is crucial for emergency response applications, where an accurate damage assessment must be achieved regardless of the quality of available prior information.

 \begin{table*}[t]
 \caption{Normalized confusion matrix for binary building damage estimation results produced by QVCBI and the baseline model\cite{rao2022earthquake} in the 2020 Zagreb earthquake.}
 \centering
 \begin{tabular}{|c|c|c|c|c|}
\toprule
Damage Grade &\multicolumn{2}{|c|}{Damaged (Estimated)
} & \multicolumn{2}{|c|}{Undamaged (Estimated)
}\\
\hline
Models & Our Model & Baseline model &  Our Model & Baseline model\\
\hline
Damaged (True) &  \textbf{0.9899 (TP)} & 0.82(TP)& \textbf{0.0101 (FP)} & 0.18 (FP)  \\
\hline
Undamaged (True) & \textbf{0.0578 (FN)} & 0.52 (FN) & \textbf{0.9422 (TN)} & 0.48 (FP)\\
\bottomrule
 \end{tabular}
 \label{conf_Zag_1}
 \end{table*}

\subsubsection{The 2019 Ridgecrest, California earthquake}

The 2019 Ridgecrest earthquake provides a critical validation of the capability of QVCBI for joint hazard estimation. Unlike previous cases that focused solely on building damage assessment, this case study demonstrates the ability of QVCBI to simultaneously estimate building damage, landslides, and liquefaction probabilities across a complex disaster scenario. As shown in Table \ref{RC_LSLFAUC}, we systematically evaluate QVCBI using three different building damage prior configurations while maintaining consistent priors for landslide and liquefaction. This experimental design enables us to assess both the robustness of QVCBI to prior model quality and its ability to maintain accurate joint hazard estimation. The results reveal several key strengths of our approach.

First, in building damage assessment, QVCBI demonstrates consistent improvements across all prior configurations. Using the HAZUS prior achieves the highest performance (AUC: 0.9671), representing a 4.77\% improvement over the prior (AUC: 0.9231). The combined prior and PAGER prior configurations also show substantial improvements, achieving AUCs of 0.9417 and 0.9316 respectively. Second, and more significantly, QVCBI maintains robust performance in secondary hazard estimation regardless of building damage prior quality. When using the HAZUS prior, the model achieves strong performance for both landslide (AUC: 0.9507) and liquefaction (AUC: 0.8645) estimation. Even with the poorly-performing PAGER prior, which has an AUC of just 0.5094 for landslides in its prior form, QVCBI maintains excellent estimation capability (LS AUC: 0.9294, LF AUC: 0.8213).

This consistent performance across all hazard types, regardless of prior model quality, validates a fundamental advantage of our causal Bayesian approach: its ability to leverage the complex interactions between different hazards through the causal network structure. The framework effectively utilizes these relationships to maintain reliable estimates even when individual prior models are weak, demonstrating its practical value for comprehensive post-earthquake hazard assessment. The results also highlight the scalability of QVCBI to multiple hazard types without compromising accuracy. The maintenance of high AUC values across all three hazard categories, even under varying prior conditions, suggests that the framework can effectively handle the increased complexity of joint hazard estimation while maintaining computational efficiency.

 \begin{table*}[t]
\caption{AUC results for joint estimation of building damage (BD), landslide (LS), and liquefaction (LF) in the Ridgecrest earthquake. The posterior results show the performance of our model using different building damage priors, while maintaining the same LS and LF priors across all scenarios.}
\begin{center}
\begin{tabular}{l|ccc}
\toprule
Model Configuration & LS & LF & BD \\
\toprule
Posterior (using the HAZUS prior as BD prior) & 0.9507 & 0.8645 & 0.9671 \\
Posterior (using the PAGER prior as BD prior) & 0.9294 & 0.8213 & 0.9316 \\
Posterior (using the combined prior as BD prior) & 0.9389 & 0.8401 & 0.9417 \\
\toprule
Prior Models: & & & \\
\toprule
HAZUS prior for BD & 0.9083 & 0.8031 & 0.9231 \\
PAGER prior for BD & 0.5094 & 0.8031 & 0.9231 \\
Combined prior for BD & 0.7379 & 0.8031 & 0.9231 \\
\toprule
\end{tabular}
\end{center}
\label{RC_LSLFAUC}
\end{table*}

\subsubsection{Computational Efficiency Analysis}
To demonstrate the suitability of our proposed framework for large-scale applications, we conducted a comprehensive evaluation of its computational efficiency across diverse earthquake scenarios. Table \ref{table:time} presents processing times for five major earthquake events with varying geographical extents. The Haiti earthquake region, our largest test area at 15,879 $km^{2}$, processes in 12,013 seconds, while the Puerto Rico region (17,012 $km^{2}$) requires 5,077 seconds. The more compact Zagreb earthquake region (6,710 $km^{2}$) processes in 8,987 seconds, with the Italy and Ridgecrest regions (approximately 11,900 $km^{2}$) completing in about 9,000 seconds each.

These results reveal interesting scaling properties of our framework. While processing time generally correlates with map size, the relationship isn't strictly linear, suggesting that other factors such as terrain complexity and building density also influence computational requirements. The framework demonstrates particularly efficient processing for the Puerto Rico region despite its larger size, likely due to more uniform terrain characteristics and building distribution patterns. The achieved processing times - ranging from 1.5 to 3.6 hours - align well with operational requirements for rapid damage assessment, particularly considering that satellite revisit times typically range from 6 hours to two days for most remote sensing platforms.

\begin{table}[h]
\centering
\caption{This table shows the time cost of running our framework using the same batch size in three earthquake events using real-world data.}
\begin{tabular}{c|ccccc}
\toprule
\textbf{Method}  & \textbf{Haiti EQ.} & \textbf{ Puerto Rico EQ.} & \textbf{Zagreb EQ.} & \textbf{Italy EQ.} & \textbf{Ridgecrest EQ.}  \\
\hline
Map size & 15,879 $km^2$ & 17,012 $km^2$ &  6,710 $km^2$ & 11,880 $km^2$ & 11,960 $km^2$ \\
Time(s) & 12,013 & 13,003 & 5,189 & 9,076 & 9,247 \\
\bottomrule
\end{tabular}%
\label{table:time}
\end{table}

\section{Discussion and conclusion}

Our quadratic variational causal Bayesian inference (QVCBI) framework addresses several critical challenges in post-disaster building damage assessment. The ability to distinguish building damage from secondary hazards while handling environmental noise represents a significant step toward more reliable rapid damage assessment after an earthquake when accurate information can save lives. The success of QVCBI in this aspect demonstrates the value of incorporating physical causal relationships into damage assessment models. The robustness of QVCBI to data quality limitations is particularly significant for global applicability. Traditional approaches often struggle in regions with limited building inventory data or poorly calibrated fragility models. The ability of QVCBI to generate reliable estimates even with incomplete building footprint data or weak prior models makes it particularly valuable for developing regions, where rapid and accurate damage assessment is often most needed but high-quality prior data is scarce.

A key advancement is the ability of QVCBI to provide multi-class damage classifications while maintaining computational efficiency at regional scales. This capability directly addresses the need for differentiated response strategies, allowing emergency managers to better prioritize resources based on damage severity. By providing more nuanced damage assessments across large areas, our approach bridges the gap between the need for detailed damage information and the practical constraints of rapid post-disaster response. While our results demonstrate significant improvements over existing approaches, future work should explore the application of the framework to other types of natural disasters and different remote sensing data sources. 

\section*{Author statement}
Xuechun Li and Susu Xu conceptualized the research, developed methodology. Xuechun Li processed the experimental data and implemented the code, conducted the experiments, analyzed the results. Both authors edited the paper and approved the submission.

\section*{Declaration of competing interest}
The authors declare that they have no known competing financial interests or personal relationships that could have appeared to influence the work reported in this paper.

\section*{Sources of funding}
The author(s) disclosed receipt of the following financial support for the research, authorship, and/ or publication of this article: X. L. and S. X. are supported by U.S. Geological Survey Grant G22AP00032 and NSF CMMI-2242590. 
Any mention of commercial products is for informational purposes and does not constitute an endorsement by the U.S. government.

\section*{Data Statement}
Data used in this study were collected from several publicly accessible sources. The primary observational data consists of Damage Proxy Maps (DPMs) generated by NASA's Advanced Rapid Imaging and Analysis (ARIA) team using InSAR imagery from Sentinel-1 satellites, available at \url{https://aria-share.jpl.nasa.gov/}. These DPMs capture correlation changes between pre- and post-event images, providing valuable information for rapid hazard and impact estimation.

For ground truth validation, we collected data from multiple sources across the three earthquake events studied. For the 2021 Haiti earthquake (M7.2), building damage and landslide inventories were provided by StEER (available at: \url{https://www.steer.network/haiti-response}) and GEER teams. Field reconnaissance data for the 2020 Puerto Rico earthquake (M6.4) was collected by USGS, University of Puerto Rico Mayagüez, GEER, and StEER teams, available at \url{https://www.sciencebase.gov/catalog/item/5eb5b9dc82ce25b5135ae83a}.

Additional data sources used in this study include USGS ShakeMap and ground failure models (\url{https://earthquake.usgs.gov/}) as prior models, along with building footprints from OpenStreetMap (\url{https://www.openstreetmap.org/}). Our model evaluation utilized both synthesized data (generated from these real-world sources) and direct real-world observations. Any data not available through these public repositories may be obtained from the corresponding author upon reasonable request.

\bibliography{main}

\appendixpage
\appendix  
\section{Example of Inaccurate Haiti BF}\label{Appendix_inaccurate_BF}
\begin{figure*}[tb]
    \centering
    \includegraphics[width=0.8\textwidth]{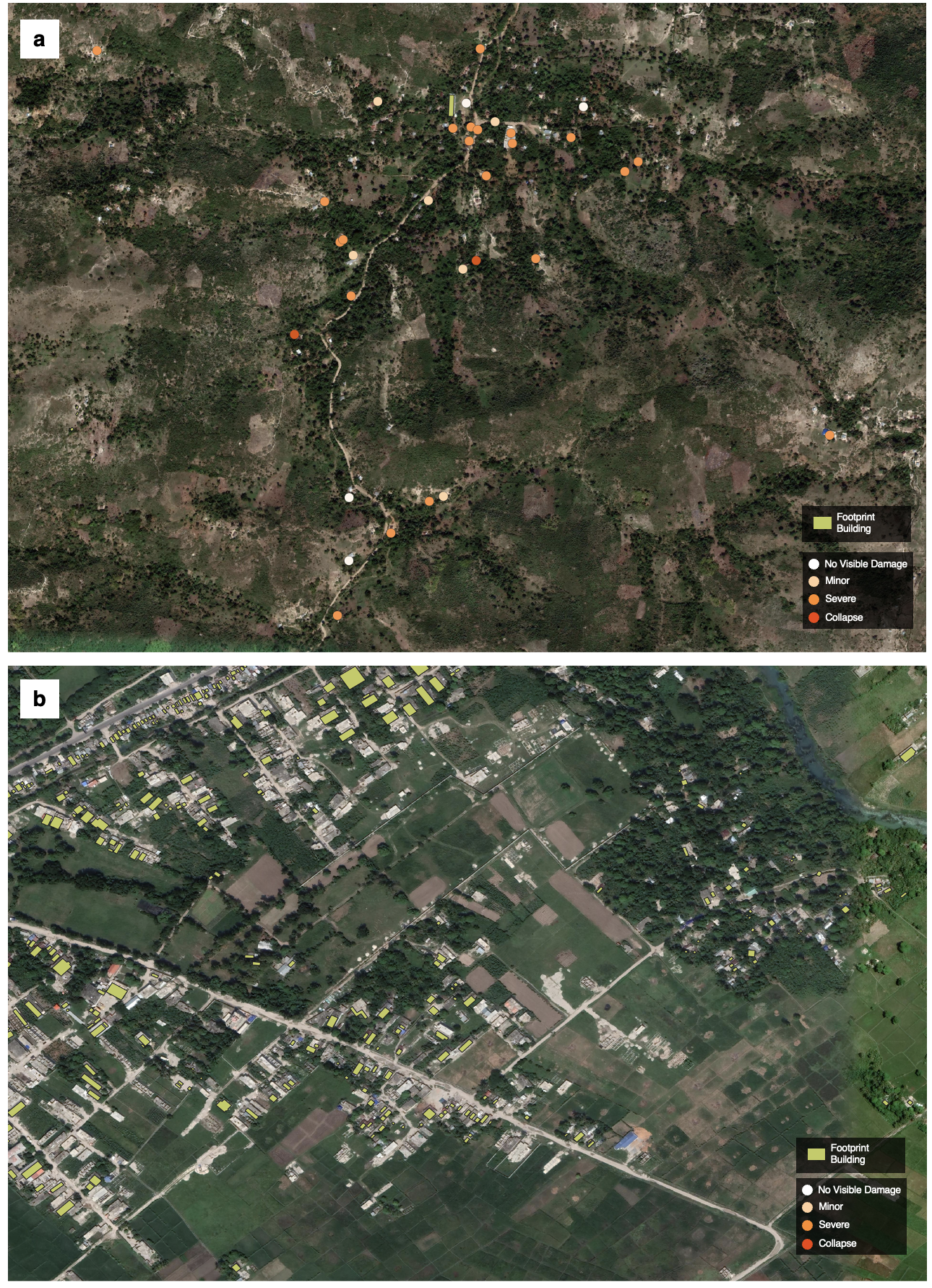}
    \caption{Satellite images with reported BF demonstrated the inaccuracy of the BF.}
    \label{inaccurateBF}
\end{figure*}

As shown in Figure \ref{inaccurateBF}, satellite images indicate areas with buildings but without reported BFs, showing the inaccuracy of the BF data in Haiti.

\section{Derivation of Variational Lower Bound}\label{Appendix_lower_bound}
\noindent We denote $\sum_{i,m_i} \mathbb{E}[\text{log}p(\epsilon_{i,m_i}^{l})] + \mathbb{E}[\text{log}p(\epsilon_{y}^{l})]$ in Equation \ref{expand} as $C_{1}$. Because $C_{1}$ is not related to our posterior and is a fixed constant, we do not need to compute the closed-form solution of it. To compute item [2]:
\begin{equation*}
\begin{aligned}
    & \mathbb{E}_{q(X^l,\epsilon^l)}[\text{log}q(X^l,\epsilon^l)] \\
    & = \underbrace{\mathbb{E}_{\epsilon_{y}^l \sim N(0,1)}[\text{log}p(\epsilon_{y}^l)] + \sum_{i}\mathbb{E}_{\epsilon_{i} \sim N(0,1)}[\text{log}p(\epsilon_{i}^l)]}_{C_{2}} \\
    & + \sum_{i, m_{i}}\mathbb{E}_{x^l_{i,m_{i}} \sim q(x^l_{i,m_{i}})}[{\text{log}\prod_{i}(q^l_{i,m_{i}})}^{I(x^l_{i} = m_{i})}] \\
    & = C_{2} + \sum_{i,m _{i}}\mathbb{E}_{x_{i,m_i} \sim q(x^l_{i,m_i})}[I(x^l_i = m_i)\text{log}q^l_{i,m_i}] \\ 
    & = C_2 + \sum_{i, m_{i}}q_{i,m_i}\text{log}q^l_{i,m_i}
\end{aligned}
\end{equation*}
\noindent where $C_1$ and $C_2$ in the items [1] and [2] cancel out. 
To estimate the item [3], we need to go back to the assumption of the conditional distribution of DPM given the parents of DPM - landslide, liquefaction, or building damage - and noise $\epsilon_{y}$. The cumulative distribution function (CDF) of the conditional distribution is:
\begin{equation}
     p(y^l|x_{\mathcal{P}(y^{l})}) = \frac{1}{y^l|w_{\epsilon_{y}}|\sqrt{2\pi}}\text{exp}[ -\frac{[\text{log}y^l - (\sum E_{m_{i}})]^2}{2w_{\epsilon_{y}}^2}]
     \label{CDF}
\end{equation}
\noindent where $E_{m_{i}}$ is in Equation \ref{Emi}. Take the logarithm of both sides of the Equation \ref{CDF}, we get:
\begin{equation}
\label{eq2}
\begin{aligned}
\text{log}p(y^l|x_{\mathcal{P}(y^l)}) & = -\text{ln}y^l - \text{ln}|w_{ \epsilon_{ y}}| - \frac{1}{2}\text{ln}2\pi - \frac{[\text{log}y^l - (\sum_{ m_{ k}}E_{m_{i}})]^2}{2w_{ \epsilon_{ y}}^2}
\end{aligned}
\end{equation}

Suppose for each $i \in \{LS, LF, BD\}$, $x_{i}$ has ($M_{i}$ + 1) categories. Then, the expectation of Equation \ref{eq2}, i.e., item [3], is:
\begin{equation*}
\begin{aligned}
& \mathbb{E}_{\begin{subarray}{c} x_{\mathcal{P}(y^l)}^l \sim q(\mathcal{P}(y^l))\\  \epsilon_{y}^l \sim \mathcal N(0,1)\end{subarray}}[\text{log}p(y^l|x_{\mathcal{P}(y^l)}^l,\epsilon_{y}^l)] \\
& = -\text{ln}y^l - \text{ln}|w_{\epsilon_{y}}| - \frac{1}{2}\text{ln}2\pi \\
& - \frac{(\text{ln}y^l)^2 + w_{0,y}^2 + w_{\epsilon_{y}}^2 + \sum_{k \in \mathcal{P}(y^l), m_{k}}w_{k,y, m_{k}}^2 m_{k}^2 q^l_{k,m_{k}}}{2w_{\epsilon_{y}}^2} \\
& - \frac{2\sum_{\begin{subarray}{c} i,j \in \mathcal{P}(y^l)\\ m_{i}, m_{j}\\ i \neq j\end{subarray}}w_{i,y, m_{i}}w_{j,y, m_{j}}(m_{i}*q^l_{i,m_{i}})(m_{j} q^l_{j,m_{j}})}{2w_{\epsilon_{y}}^2 } \\
& - \frac{2(w_{0,y}-\text{ln}y^l)\sum_{k \in \mathcal{P}(y^l), m_{k}}w_{k,y, m_{k}}m_{k}q_{k, m_{k}} - 2w_{ 0,y}\text{ln}y^l}{2w_{\epsilon_{y}}^2 }
\end{aligned}
\end{equation*}
Substitute Equation.\ref{alpha_hat} into Equation \ref{bound}, we get:
\begin{equation*}
\begin{aligned}
    \text{log}(\sum_{m_{i} = 0}^{M_{i}}e^{z_{m_{i}}})
    & \leq -\hat{\alpha}^{2}_{i}\sum_{m_{i} = 0}^{M_{i}}\lambda(\xi_{m_{i}}) +\sum_{m_{i} = 0}^{M_{i}}\text{log}(1+e^{\xi_{m_{i}}})\\
    & +\sum_{m_{i} = 0}^{M_{i}}\lambda(\xi_{m_{i}})(z_{m_{i}}^{2}-\xi_{m_{i}}^{2})
\end{aligned}
\end{equation*}
\noindent where 
\begin{equation*}
    \begin{aligned}
        \hat{\alpha}^{2}_{i} & = \frac{16\sum_{m_{i}}\lambda(\xi_{m_{i}})^{2}z_{m_{i}}^{2} + 16\sum \lambda(\xi_{m_{r}})\lambda(\xi_{m_{s}})z_{m_{r}}z_{m_{s}}}{16(\sum_{m_{i} = 0}^{M_{i}}\lambda(\xi_{m_{i}}))^{2}} \\
        & + \frac{(M_{i} - 1)^{2} + 8(M_{i} - 1)\sum_{m_{i} = 0}^{M_{i}}\lambda(\xi_{m_{i}})z_{m_{i}} }{16(\sum_{m_{i} = 0}^{M_{i}}\lambda(\xi_{m_{i}}))^{2}} 
    \end{aligned}
\end{equation*}
In our case:
\begin{equation*}
\begin{aligned}
       \text{log}p(x_{i, m_{i}}|x_{\mathcal{P}(i), m_{\mathcal{P}(i)}}, \epsilon_{i}) & = z_{m_{i}} - \text{log}(\sum_{m_{i}} z_{m_{i}}) \\
       & \geq z_{m_{i}} + \hat{\alpha_{i}}^{2}\sum_{m_{i} = 0}^{M_{i}}\lambda(\xi_{m_{i}}) \\
       & -\sum_{m_{i} = 0}^{M_{i}}\text{log}(1+e^{\xi_{m_{i}}}) \\
       & -\sum_{m_{i} = 0}^{M_{i}}\lambda(\xi_{m_{i}})(z_{m_{i}}^{2}-\xi_{m_{i}}^{2}) \\
       & -\frac{1}{2}\sum_{m_{i} = 0}^{M_{i}}(z_{m_{i}} - \xi_{m_{i}})
\end{aligned}
\end{equation*}
\noindent Therefore, we obtain the lower bound for item [4] as: 
\begin{equation*}
\begin{aligned}
    & \sum_{i, m_{i}}\mathbb{E}_{\begin{subarray}{c} x_{i,m_{i}} \sim q(x_{i,m_{i}})\\ x_{\mathcal{P}(i)} \sim q(x_{\mathcal{P}(i),m_{\mathcal{P}(i}})\\ \epsilon_{i,m_{i}} \sim N(0,1) \end{subarray}}[\text{log}p(x_{i,m_{i}}|\epsilon_{i},x_{\mathcal{P}(i)})] \\
    & \geq  \sum_{i, m_{i}}m_{i}q_{i, m_{i}}[ \mathbb{E}(z_{m_{i}}) + \mathbb{E}(\hat{\alpha}^{2})\sum_{m_{i} = 0}^{M_{i}}\lambda(\xi_{m_{i}})\\
    &-\sum_{m_{i} = 0}^{M_{i}}\text{log}(1+e^{\xi_{m_{i}}})  -\sum_{m_{i} = 0}^{M_{i}}\lambda(\xi_{m_{i}})(\mathbb{E}(z_{m_{i}}^{2})-\xi_{m_{i}}^{2})\\
    &-\frac{1}{2}\sum_{m_{i} = 0}^{M_{i}}(\mathbb{E}(z_{m_{i}}) - \xi_{m_{i}})]
\end{aligned} 
\end{equation*}
\noindent where:
\begin{equation*}
    \mathbb{E}(z_{m_{i}})  =  w_{0,i, m_{i}}  + \sum_{k \in {\mathcal{P}(i^l)}, m_{k}}w_{k, i, m_{i}}m_{k}q_{k, m_{k}}
\end{equation*}
\begin{equation*}
\begin{aligned}
    \mathbb{E}(z_{m_{i}}^{2}) = &
    \sum_{k \in {\mathcal{P}(i^l)}, m_{k}}w_{k, i, m_{i}}^{2}m_{k}^{2}q_{k, m_{k}} + w_{\epsilon_{i}, m_{i}}^{2} + w_{0,i, m_{i}}^{2} \\
    & + \sum_{\begin{subarray}{c}r, s \in \mathcal{P}(i)\\ r \neq s \\ m_{r}, m_{s} \end{subarray}}w_{r, i, m_{i}}w_{s, i, m_{i}}m_{r}m_{s}q_{r, m_{r}}q_{s, m_{s}} \\
    & + 2w_{0,i, m_{i}}\sum_{k \in {\mathcal{P}(i^l)}, m_{k}}w_{k, i, m_{i}}m_{k}q_{k, m_{k}}
\end{aligned}
\end{equation*}
\begin{equation*}
\begin{aligned}
\mathbb{E}(z_{m_{r}}z_{m_{s}}) & = w_{0, r, m_{r}}w_{0, s, m_{s}}
    \\
    & + w_{0, r, m_{r}}\sum_{k \in \mathcal{P}(r), m_{k}}w_{k, r, m_{r}}m_{k}q_{k, m_{k}}\\
    & + w_{0, s, m_{s}} \sum_{l \in \mathcal{P}(s)}w_{l, s, m_{s}}m_{l}q_{l, m_{l}} \\ 
    & + \sum_{k, l,m_{k}, m_{l}}w_{k, r, m_{r}}w_{l, s, m_{s}}m_{k}m_{l}q_{k, m_{k}}q_{l, m_{l}}
\end{aligned}
\end{equation*}
\begin{equation*}
\begin{aligned}
    \mathbb{E}(\hat{\alpha}^{2}) & = \frac{16\sum_{m_{i} = 0}^{M_{i}}\lambda(\xi_{m_{i}})^{2}\mathbb{E}(z_{m_{i}}^{2})
    + (M_{i} - 1)^{2}}{16(\sum_{m_{i} = 0}^{M_{i}}\lambda(\xi_{m_{i}}))^{2}} \\
    & + \frac{16\sum_{r \neq s} \lambda(\xi_{m_{r}})\lambda(\xi_{m_{s}})\mathbb{E}(z_{m_{r}}z_{m_{s}})}{16(\sum_{m_{i} = 0}^{M_{i}}\lambda(\xi_{m_{i}}))^{2}} +  \frac{8(M_{i} - 1)\sum_{m_{i} = 0}^{M_{i}}\lambda(\xi_{m_{i}})\mathbb{E}(z_{m_{i}})}{16(\sum_{m_{i} = 0}^{M_{i}}\lambda(\xi_{m_{i}}))^{2}}
\end{aligned}
\end{equation*}
\begin{equation*}
\begin{aligned}
\mathbb{E}(z_{m_{r}}z_{m_{s}}) & = w_{0, r, m_{r}}w_{0, s, m_{s}}\\
    & + w_{0, r, m_{r}}\sum_{k \in \mathcal{P}(r), m_{k}}w_{k, r, m_{r}}m_{k}q_{k, m_{k}} \\
    & + w_{0, s, m_{s}} \sum_{l \in \mathcal{P}(s)}w_{l, s, m_{s}}m_{l}q_{l, m_{l}} \\ 
    & + \sum_{k, l,m_{k}, m_{l}}w_{k, r, m_{r}}w_{l, s, m_{s}}m_{k}m_{l}q_{k, m_{k}}q_{l, m_{l}}
\end{aligned}
\end{equation*}

The item [5] is the expectation of the mutually exclusive variable. Following the definition of the variable, for those locations $l$ where the mutual exclusively between landslide and liquefaction is ascertained, there should be $u^{l} = 0$, so we have:
\begin{equation*}
\begin{aligned}
    \mathbb{E}_{q(X_{\mathcal{P}(u^l)})}[\text{log}p(u^l|x^{l}_{\mathcal{P}(u),m_{\mathcal{P}(u^l)}})] & =  -\frac{1}{2}\text{log}2\pi \sigma^2 - \frac{\prod_{k}\sum_{m_{k}}m_{k}^2 q^l_{k,m_{k}}}{2\sigma^2}
\end{aligned}
\end{equation*}

Given a map containing a set of locations, $L$, we further derive a tight lower bound for the log-likelihood as follows:
\begin{equation*}
\begin{aligned}
    & \mathcal{L}(\textbf{q,w})  = \text{log}P(Y,U) = \sum_{l \in \mathcal{L}}\text{log}P(y^{l},u^{l}) \\
    & \geq \sum_{l \in \mathcal{L}} \{-\text{ln}y^{l} - \text{ln}|w_{\epsilon_{y}}| - \frac{1}{2}\text{ln}2\pi \\
    & - \frac{(\text{ln}y^{l})^2 + w_{0,y}^2 + w_{\epsilon_{y}}^2 + \sum_{k, m_{k}}w_{k,y,m_{k}}^2 m_{k}^2 q_{k,m_{k}}}{2w_{\epsilon_{y}}^2}\\
    & - \frac{\sum_{\begin{subarray}{c} i,j \in \mathcal{P}(y^l)\\ m_{i}, m_{j}\\ i \neq j\end{subarray}}w_{i,y,m_{i}}w_{j,y,m_{j}}(m_{i}q^l_{i,m_{i}})(m_{j} q^l_{j,m_{j}})}{w_{\epsilon_{y}}^2} \\ 
    & - \frac{(w_{0,y}-\text{ln}y^l)\sum_{k, m_{k}}w_{k,y,m_{k}} m_{k}q_{k, m_{k}}- w_{0,y}\text{ln}y^l}{w_{\epsilon_{y}}^2} \\
    & + \sum_{i, m_{i}}m_{i}q_{i, m_{i}}[ \mathbb{E}(z_{m_{i}}) + \mathbb{E}(\hat{\alpha}^{2})\sum_{m_{i} = 0}^{M_{i}}\lambda(\xi_{m_{i}})\\
    & -\sum_{m_{i} = 0}^{M_{i}}\text{log}(1+e^{\xi_{m_{i}}}) - \sum_{m_{i} = 0}^{M_{i}}\lambda(\xi_{m_{i}})(\mathbb{E}(z_{m_{i}}^{2})-\xi_{m_{i}}^{2})]
\end{aligned}
\end{equation*}
\begin{equation*}
\begin{aligned}
    & -\sum_{i, m_{i}}\sum_{m_{i} = 0}^{M_{i}}\frac{1}{2}m_{i}q_{i, m_{i}}(\mathbb{E}(z_{m_{i}}) - \xi_{m_{i}}) -\frac{1}{2}\text{log}2\pi \sigma^2 \\
    & - \frac{\prod_{k\in \mathcal{P}(u^l)}\sum_{m_{k} = 0}^{M_{k}}m_{k}^2 q^l_{k,m_{k}}}{2\sigma^2}  - \sum_{i,m_{i}}q_{i,m_i}\text{log}q^l_{i,m_i}\}
\end{aligned}
\end{equation*}

\section{Derivation of Expectation-Maximization Algorithm}\label{appendix_EM}
We define $\mathcal{T}(\cdot)$ in Equation \ref{EM} as follows: 
\begin{equation*}
\begin{aligned}
    \mathcal{T}(\cdot) & = - \frac{w_{k,y,m_{k}}^{2}m_{k}^{2}}{2w_{\epsilon_{y}}^{2}} - \frac{(w_{0,y} - \text{ln}y)w_{k,y,m_{k}}m_{k}}{2w_{\epsilon_{y}}^{2}}  \\
    & - \frac{(\sum_{j \in S(k,y), m_{j}}w_{j,y, m_{j}}m_{j}q_{j, m_{j}})w_{k,y,m_{k}}m_{k}}{2w_{\epsilon_{y}}^{2}} \\
    & + m_{k}[ \mathbb{E}(z_{m_{k}}) + \mathbb{E}(\hat{\alpha}^{2})\sum_{m_{k} = 0}^{M_{k}}\lambda(\xi_{m_{k}}) \\
    & -\sum_{m_{k} = 0}^{M_{k}}\text{log}(1+e^{\xi_{m_{k}}}) 
     - \frac{m_{k}}{2}\sum_{m_{k}}(\mathbb{E}(z_{m_{k}}) - \xi_{m_{k}})\\
    &  - \sum_{m_{k} = 0}^{M_{k}}\lambda(\xi_{m_{k}})(\mathbb{E}(z_{m_{k}}^{2})-\xi_{m_{k}}^{2}) \\
    & - \frac{\prod_{i \in S(k,u)}m_{k}^{2}\sum_{m_{i} = 0}^{M_{\scriptscriptstyle} i}m_{i}^{2}q_{i, m_{i}}}{2\sigma^{2}}\\
    & + \sum_{i \in C(k), m_{i}}m_{i}q_{i, m_{i}}[\frac{\partial \mathbb{E}(z_{m_{i}})}{\partial q_{k, m_{k}}} + \frac{\partial \mathbb{E}(\hat{\alpha}^{2})}{\partial q_{k, m_{k}}}\sum_{m_{i}}\lambda(\xi_{m_{i}}) \\
    & - \sum_{m_{i}}\lambda(\xi_{m_{i}})\frac{\partial \mathbb{E}(z_{m_{i}}^{2})}{\partial q_{k, m_{k}}} - \frac{1}{2}\sum_{m_{i}}\frac{\partial \mathbb{E}(z_{m_{i}})}{\partial q_{k, m_{k}}}]
\end{aligned}
\end{equation*}

The partial derivatives in $\mathcal{T}(\cdot)$ are:
\begin{eqnarray*}
&& \frac{\partial  \mathbb{E}(z_{m_{i}})}{\partial q_{k, m_{k}}}  = w_{k, i, m_{i}}m_{k} \\
&&
 \begin{aligned}
 \frac{\partial \mathbb{E}(z_{m_{i}}^{2})}{\partial q_{k, m_{k}}} & = w_{k, i, m_{i}}^{2}m_{k}^{2} \\
 & + \sum_{r \in S(k,i)}w_{k, i, m_{i}}m_{k}w_{r, i, m_{i}}m_{r}q_{r, m_{r}} \\
& + 2w_{0, i, m_{i}}w_{k, i, m_{i}}m_{k}
\end{aligned}
\end{eqnarray*}
\begin{equation*}
\begin{aligned}
 \frac{\partial \mathbb{E}(z_{m_{r}}z_{m_{s}})}{\partial q_{k, m_{k}}} & = w_{0, r, m_{r}}w_{k, r, m_{r}}m_{k} \\
 & + \sum_{l \in \mathcal{P}(S)}w_{k, r, m_{r}}m_{k}w_{l, s, m_{s}}m_{l}q_{l, m_{l}}, \\
 & r \in \mathcal{C}(k), l \notin \mathcal{C}(k) 
 \end{aligned}
\end{equation*}
\begin{equation*}
\begin{aligned}
 \frac{\partial\mathbb{E}(\hat{\alpha}^{2})}{\partial q_{k, m_{k}}} & = \frac{\sum_{m_{i} = 0}^{M_{i}}\lambda(\xi_{m_{i}})^{2}\frac{\partial \mathbb{E}(z_{m_{i}}^{2})}{\partial q_{k, m_{k}}}}{(\sum_{m_{i} = 0}^{M_{i}}\lambda(\xi_{m_{i}}))^{2}}\\
 &+\frac{\sum_{r \neq s}\lambda(\xi_{m_{r}})\lambda(\xi_{m_{s}})\frac{\partial \mathbb{E}(z_{m_{r}}z_{m_{s}})}{\partial q_{k, m_{k}}}}{(\sum_{m_{i} = 0}^{M_{i}}\lambda(\xi_{m_{i}}))^{2}}\\
  &+\frac{(M_{i} - 1)\sum_{m_{i} = 0}^{M_{i}}\lambda(\xi_{m_{i}})\frac{\partial \mathbb{E}(z_{m_{i}})}{\partial q_{k, m_{k}}}}{2(\sum_{m_{i} = 0}^{M_{i}}\lambda(\xi_{m_{i}}))^{2}}
\end{aligned}
\end{equation*}
The partial derivative of the log-likelihood with respect to the weights between latent variables nodes as: 
\begin{equation*}
\begin{aligned}
\frac{\partial \mathcal{L}}{\partial w_{k, i, m_{i}}} & = \frac{\partial \mathbb{E}(z_{m_{i}})}{\partial w_{k, i, m_{i}}} + \frac{\partial \mathbb{E}(\hat{\alpha})^{2}}{\partial w_{k, i, m_{i}}}\sum_{m_{i}=0}^{M_{i}}\lambda(\xi_{m_{i}}) \\
& - \sum_{m_{i}=0}^{M_{i}}\lambda(\xi_{m_{i}})\frac{\partial \mathbb{E}(z_{m_{i}}^{2})}{\partial w_{k, i, m_{i}}} 
- \frac{1}{2}\sum_{m_{i}=0}^{M_{i}}\frac{\partial \mathbb{E}(z_{m_{i}})}{\partial w_{k, i, m_{i}}} 
 \end{aligned}
\end{equation*}
\begin{equation*}
\frac{\partial \mathcal{L}}{\partial w_{\epsilon_{i}, m_{i}}}
= \frac{2\sum_{m_{i} = 0}^{M_{i}}\lambda(\xi_{i})^{2}w_{\epsilon_{i}, m_{i}}}{\lambda(\xi_{i})} - 2\sum_{m_{i} = 0}^{M_{i}}\lambda(\xi_{i})w_{\epsilon_{i}, m_{i}}
\end{equation*}
\noindent where for $i \in \mathcal{C}(k)$: 
\begin{eqnarray*}
&& \frac{\partial \mathbb{E}(z_{m_{i}})}{\partial w_{k, i, m_{i}}}  = m_{k}q_{k, m_{k}}\\
&& 
\begin{aligned}
\frac{\partial \mathbb{E}(z_{m_{i}}z_{m_{s}})}{\partial w_{k, i, m_{i}}} & = w_{0,i,m_{i}}m_{k}q_{k, m_{k}} \\
& + m_{k}q_{k, m_{k}}\sum_{l \in \mathcal{P}(S)}w_{l,s,m_{s}}m_{l}q_{l,m_{l}}
\end{aligned}
\end{eqnarray*}
\begin{eqnarray*}
&&
\begin{aligned}
 \frac{\partial \mathbb{E}(z_{m_{i}}^{2})}{\partial w_{k, i, m_{i}}} & = 2m_{k}^{2}q_{k, m_{k}}w_{k,i,m_{i}} + 2m_{k}q_{k, m_{k}}w_{0,i,m_{i}} \\
 & + m_{k}q_{k, m_{k}}\sum_{r \in S(k,i)}w_{r, i, m_{i}}m_{r}q_{r, m_{r}}
 \end{aligned}
\\
&&
\begin{aligned}
\frac{\partial \mathcal{L}}{\partial w_{k, i, m_{i}}} & 
= \frac{\partial \mathbb{E}(z_{m_{i}})}{\partial w_{k, i, m_{i}}} + \frac{\partial \mathbb{E}(\hat{\alpha})^{2}}{\partial w_{k, i, m_{i}}}\sum_{m_{i}=0}^{M_{i}}\lambda(\xi_{m_{i}}) \\
& - \sum_{m_{i}=0}^{M_{i}}\lambda(\xi_{m_{i}})\frac{\partial \mathbb{E}(z_{m_{i}}^{2})}{\partial w_{k, i, m_{i}}} 
- \frac{1}{2}\sum_{m_{i}=0}^{M_{i}}\frac{\partial \mathbb{E}(z_{m_{i}})}{\partial w_{k, i, m_{i}}}
\end{aligned}
\end{eqnarray*}
We also have the gradient of weights between DPM and its parent nodes as follows:
\begin{eqnarray*}
&& 
\begin{aligned}
\frac{\partial \mathcal{L}}{\partial w_{k,y,m_{k}}} & = 
-\frac{(\sum_{j \in S(k,y), m_{j}}w_{j, y, m_{j}}m_{j}q_{j, m_{j}})m_{k}q_{k, m_{k}}}{w_{\epsilon_{y}}^{2}}\\
& -\frac{(w_{k,y,m_{k}}m_{k} + w_{0,y} - \text{ln}y)m_{k}q_{k, m_{k}}}{w_{\epsilon_{y}}^{2}}
\end{aligned}
\\
&& 
\begin{aligned}
    \frac{\partial \mathcal{L}}{\partial w_{\epsilon,y}} & = - \frac{1}{w_{\epsilon_{y}}} + 
    \frac{1}{w_{\epsilon_{y}}^{3}}[(\text{ln}y^l)^2 + w_{0,y}^2] \\
    &+ \frac{1}{w_{\epsilon_{y}}^{3}}[\sum_{k \in \mathcal{P}(y^l), m_{k}}w_{k,y,m_{k}}^{2} m_{k}^2 q^l_{k,m_{k}} - 2w_{0,y}\text{ln}y^l] \\
    & + \frac{1}{w_{\epsilon_{y}}^{3}}[2\sum_{\begin{subarray}{c} i,j \in \mathcal{P}(y^l)\\ m_{i}, m_{j}\\ i \neq j\end{subarray}}w_{i,y,m_{i}}w_{j,y,m_{j}}m_{i}q^l_{i,m_{i}}m_{j} q^l_{j,m_{j}} \\
    &+ \frac{1}{w_{\epsilon_{y}}^{3}}[2(w_{0,y}-\text{ln}y^l)\sum_{k, m_{k}}w_{k,y,m_{k}}m_{k}q_{k,m_{k}}]
\end{aligned}
\end{eqnarray*}
The gradients of weights between the leak node to X and y are:
\begin{equation*}
    \frac{\partial \mathcal{L}}{\partial w_{0,y}} = \frac{\text{ln}y - \sum_{k \in \mathcal{P}(y), m_{k}}w_{k,y,m_{k}}m_{k}q_{k, m_{k}} - w_{0,y}}{w_{\epsilon_{y}}^{2}}
\end{equation*}
\begin{eqnarray*}
&&
\begin{aligned}
\frac{\partial \mathcal{L}}{\partial w_{0, i, m_{i}}} & = \frac{\partial \mathbb{E}(z_{m_{i}})}{\partial w_{0, i, m_{i}}} + \frac{\partial \mathbb{E}(\hat{\alpha})^{2}}{\partial w_{0, i, m_{i}}}\sum_{m_{i}=0}^{M_{i}}\lambda(\xi_{m_{i}}) \\
& - \sum_{m_{i}=0}^{M_{i}}\lambda(\xi_
{m_{i}})\frac{\partial \mathbb{E}(z_{m_{i}}^{2})}{\partial w_{0, i, m_{i}}} 
- \frac{1}{2}\sum_{m_{i}=0}^{M_{i}}\frac{\partial \mathbb{E}(z_{m_{i}})}{\partial w_{0, i, m_{i}}}
\end{aligned}
\end{eqnarray*}
\noindent where for $i \in \mathcal{C}(k)$:
\begin{eqnarray*}
\end{eqnarray*}
\begin{eqnarray*}
&& \frac{\partial \mathbb{E}(z_{m_{i}})}{\partial w_{0, i, m_{i}}} = 1 \\
&& \frac{\partial \mathbb{E}(z_{m_{i}}^{2})}{\partial w_{0, i, m_{i}}} =
2w_{0, i, m_{i}} + 2\sum_{k \in \mathcal{P}(i)}w_{k, i, m_{i}}m_{k}q_{k, m_{k}} \\
&& 
\begin{aligned}
\frac{\partial \mathbb{E}(z_{m_{i}}z_{m_{s}})}{\partial w_{0, i, m_{i}}} & = w_{0,s,m_{s}} + \sum_{k \in \mathcal{P}(i)}w_{k, i, m_{i}}m_{k}q_{k, m_{k}}, \\
& s \neq i 
\end{aligned}
\\ 
&& 
\begin{aligned}
\frac{\partial \mathbb{E}(\hat{\alpha})^{2}}{\partial w_{0, i, m_{i}}} & = \frac{\sum_{m_{i} = 0}^{M_{i}}\lambda(\xi_{m_{i}})^{2}\frac{\partial \mathbb{E}(z_{m_{i}}^{2})}{\partial w_{0,i,m_{i}}}}{(\sum_{m_{i} = 0}^{M_{i}}\lambda(\xi_{m_{i}}))^{2}} \\
& + \frac{\sum_{r \neq s}\lambda(\xi_{m_{r}})\lambda(\xi_{m_{s}})
\frac{\partial \mathbb{E}(z_{m_{r}}z_{m_{s}})}{\partial w_{0,i,m_{i}}}}{(\sum_{m_{i} = 0}^{M_{i}}\lambda(\xi_{m_{i}}))^{2}}\\
& + \frac{(M_{i} - 2)\sum_{m_{i} = 0}^{M_{i}}
\lambda(\xi_{m_{i}})
\frac{\partial \mathbb{E}(z_{m_{i}})}{\partial w_{0,i,m_{i}}}
}{2(\sum_{m_{i} = 0}^{M_{i}}\lambda(\xi_{m_{i}}))^{2}}
\end{aligned}
\end{eqnarray*}
We also need to update $\xi_{m_{i}}$ in each iteration. Define: 
\begin{eqnarray*}
&& g(\xi_{m_{i}}) = 16(\sum_{m_{i} = 0}^{M_{i}}\lambda(\xi_{m_{i}}))^{2} \\
&&
\begin{aligned}
f(\xi_{m_{i}}) & =  16\sum_{m_{i} = 0}^{M_{i}}\lambda(\xi_{m_{i}})^{2}\mathbb{E}(z_{m_{i}}^{2}) \\
& + 16\sum_{r \neq s} \lambda(\xi_{m_{r}})\lambda(\xi_{m_{s}})\mathbb{E}(z_{m_{r}}z_{m_{s}}) \\
& + (M_{i} - 1)^{2} 
+ 8(M_{i} - 1)\sum_{m_{i} = 0}^{M_{i}}\lambda(\xi_{m_{i}})\mathbb{E}(z_{m_{i}})
\end{aligned}
\end{eqnarray*}
The gradients with respect to $\xi_{m_{i}}$ are:
\begin{eqnarray*}
&& \frac{\partial \mathbb{E}(\hat{\alpha}^{2})}{\partial \xi_{m_{i}}} = \frac{f'(\xi_{m_{i}})g(\xi_{m_{i}}) - f(\xi_{m_{i}})g'(\xi_{m_{i}})}{(g(\xi_{m_{i}}))^{2}}\\
&& 
\begin{aligned}
\frac{\partial \lambda(\xi_{m_{i}})}{\partial \xi_{m_{i}}} & = -\frac{1}{2\xi_{m_{i}}}(\frac{1}{1 + e^{-\xi_{m_{i}}}} + \frac{1}{2})\\
& + \frac{1}{2\xi_{m_{i}}e^{\xi_{m_{i}}}(1+e^{-\xi_{m_{i}}})^{2}}
\end{aligned}
\end{eqnarray*}
\begin{equation*}
\begin{aligned}
\frac{\partial \mathcal{L}}{\partial \xi_{m_{i}}} & = -\frac{e^{\xi_{m_{i}}}}{1 + e^{\xi_{m_{i}}}} - \mathbb{E}(z_{m_{i}}^{2})\frac{\partial \lambda(\xi_{m_{i}})}{\partial \xi_{m_{i}}} + 2\xi_{m_{i}}\lambda(\xi_{m_{i}}) \\
& + \xi_{m_{i}}^{2}\frac{\partial \lambda(\xi_{m_{i}})}{\partial \xi_{m_{i}}}    + \frac{\partial \mathbb{E}(\hat{\alpha}^{2})}{\partial \xi_{m_{i}}}\sum_{m_{i} = 0}^{M_{i}}\lambda(\xi_{m_{i}})\\
& + \mathbb{E}(\hat{\alpha}^{2})\frac{\partial \lambda(\xi_{m_{i}})}{\partial \xi_{m_{i}}}
\end{aligned}
\end{equation*}
\noindent where:
\begin{equation*}
\frac{\partial g(\xi_{m_{i}})}{\partial \xi_{m_{i}}} = 32\frac{\partial \lambda(\xi_{m_{i}})}{\partial \xi_{m_{i}}}\sum_{m_{i} = 0}^{M_{i}}\lambda(\xi_{m_{i}})
\end{equation*}
\begin{equation*}
  \begin{aligned}
\frac{\partial f(\xi_{m_{i}})}{\partial \xi_{m_{i}}} & = 32\lambda(\xi_{m_{i}})\frac{\partial \lambda(\xi_{m_{i}})}{\partial \xi_{m_{i}}}\mathbb{E}(z_{m_{m_{i}}}^{2}) \\
& + 16\sum_{r \neq i}\frac{\partial \lambda(\xi_{m_{i}})}{\partial \xi_{m_{i}}}\lambda(\xi_{m_{r}})\mathbb{E}(z_{m_{i}}z_{m_{r}}) \\
& + 8(M_{i} - 1)\mathbb{E}(z_{m_{i}})\frac{\partial \lambda(\xi_{m_{i}})}{\partial \xi_{m_{i}}}
\end{aligned}
\end{equation*}

\end{document}